\documentclass[journal]{IEEEtran}

\usepackage[noend]{algpseudocode}
\usepackage{algorithmicx,algorithm}
\usepackage{bm}
\usepackage[usenames]{color}

\usepackage{graphicx}
\usepackage{tabularx}
\usepackage{amsmath}
\usepackage{booktabs}
\usepackage{amssymb}
\usepackage{cite}

\usepackage[caption=false,font=footnotesize]{subfig}

\ifCLASSINFOpdf
\else
\fi
\begin{document}
	
	%
	\title{Gaussian Dynamic Convolution for Efficient Single-Image Segmentation}
	%
	%
	%
	
	\author{Xin~Sun,~\IEEEmembership{Member,~IEEE,}
		Changrui~Chen,	
		Xiaorui~Wang,	
		Junyu~Dong,~\IEEEmembership{Member,~IEEE,}\\
		Huiyu~Zhou,		
		Sheng~Chen,~\IEEEmembership{Fellow,~IEEE}
		
	\thanks{This work was supported in part by National Natural Science Foundation of China under Project Nos. 61971388, U1706218, and L1824025, and European Union’s Horizon 2020 research and
		innovation program under the Marie-Sklodowska-Curie grant agreement No. 720325.} 
		
	\thanks{X. Sun, X. Wang and J. Dong are with Department of Computer Science and Technology, Ocean University of China, Qingdao, Shandong Province, 266100 China. X Sun is also with the Department of Aerospace and Geodesy, Technical University of Munich, Germany. (e-mails: sunxin1984@ieee.org,  recyclerblacat@stu.ouc.edu.cn, dongjunyu@ouc.edu.cn.)} %
	\thanks{C. Chen is with WMG Data Science, University of Warwick, Coventry, CV4 7AL, United Kingdom (e-mail: geoffreychen777@gmail.com).}%
	\thanks{H Zhou is with Department of Informatics, University of Leicester, LE1 7RH, UK (e-mail: hz143@leicester.ac.uk).} %
	\thanks{S.~Chen is with School of Electronics and Computer Science, University of Southampton, Southampton SO17 1BJ, UK (e-mail: sqc@ecs.soton.ac.uk).} %
	\vspace*{-5mm}
}

	\maketitle
	

\begin{abstract}
Interactive single-image segmentation is ubiquitous in the scientific and commercial imaging software. In this work, we focus on the single-image segmentation problem only with some seeds such as scribbles. Inspired by the dynamic receptive field in the human being's visual system, we propose the Gaussian dynamic convolution (GDC) to fast and efficiently aggregate the contextual information for neural networks. The core idea is randomly selecting the spatial sampling area according to the Gaussian distribution offsets. Our GDC can be easily used as a module to build lightweight or complex segmentation networks. We adopt the proposed GDC to address the typical single-image segmentation tasks. Furthermore, we also build a Gaussian dynamic pyramid Pooling to show its potential and generality in common semantic segmentation. Experiments demonstrate that the GDC outperforms other existing convolutions on three benchmark segmentation datasets including Pascal-Context, Pascal-VOC 2012, and Cityscapes. Additional experiments are also conducted to illustrate that the GDC can produce richer and more vivid features compared with other convolutions. In general, our GDC is conducive to the convolutional neural networks to form an overall impression of the image. 
\end{abstract}


%
\IEEEpeerreviewmaketitle

\section{Introduction}
%
%
%
%
\IEEEPARstart
{S}{emantic} segmentation aims to compute a dense label prediction for each pixel in an image. It is used ubiquitously across all scientific and commercial fields where imaging has become the most critical step \cite{Yu2021}. Recent success of semantic segmentation lies on the end-to-end training with large-scale segmentation annotations. In scientific and commercial software, however, interactive image segmentation from a single image is recognized as a most user-friendly and practical operation. For example, Quick Selection tool in Adobe Photoshop is a typical commercial implementation of scribble-based interactive segmentation method. It does not rely on a large training dataset. Instead, it needs to be optimized on each image independently. Such single-image segmentation task is vital in interactive segmentation \cite{Boykov:1999ug, Tang:2013jo, 9069887} and weakly supervised segmentation. This work proposes the Gaussian dynamic convolution (GDC) to build a lightweight convolution neural network for fast and effectively accomplishing the single-image segmentation task. The network can be optimized in a flash only with one image and some seeds such as the scribbles. The proposed GDC can be easily integrated into various convolution modules. The background and motivation of our proposed GDC are now further elaborated.

The essence of semantic segmentation is to identify distinctive features for different categories. The receptive field is of great significance for the segmentation task \cite{2015arXiv151107122Y, 2016arXiv161201105Z}. Deep features with different receptive fields represent various levels of the visual attributes \cite{2015arXiv151107122Y}. It is informative to first revisit the way that human beings observe the visual world. Undoubtedly, there is a receptive field mechanism in the human visual system. The question is that \textit{do human beings adopt a fixed scale or a group of fixed scales of receptive fields when they observe objects?} The receptive field in the human being's visual system is totally dynamic \cite{gilbert1992receptive}. During the long period of growing up from a baby to an adult, the dynamic receptive field provides us stereoscopic and vivid information about the world \cite{daw2006visual}. What we store in our minds are living objects rather than some rigid features in several scales. Therefore, it is far better to equip the neural networks with stochastic dynamical receptive field for flexibly capturing context. To this end, we propose the GDC, a novel convolution kernel with a dynamic receptive field, to extract richer features for segmentation tasks.

\begin{figure}[b]
\vspace*{-6mm}
	\begin{center}
		\captionsetup[subfigure]{}
		\subfloat[]{
			\begin{minipage}[t]{0.4\columnwidth}
				\centering
				\includegraphics[width=1.2in,height=1.2in]{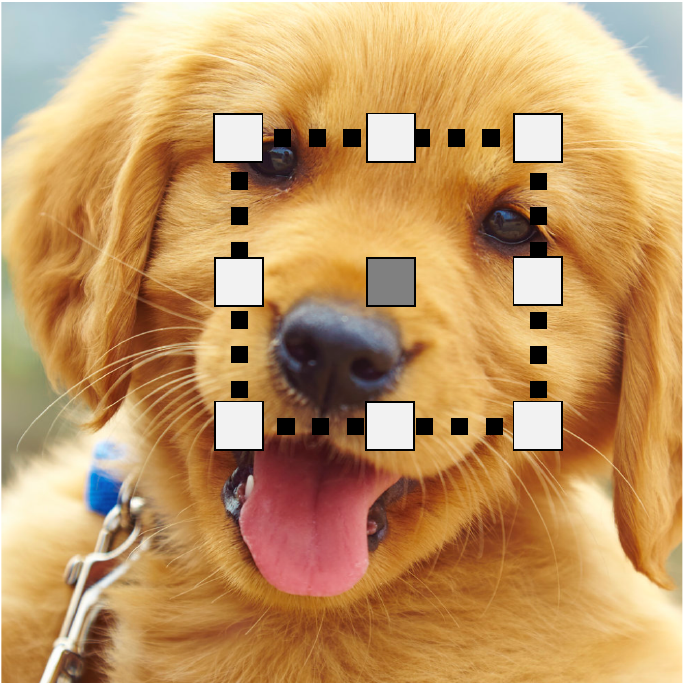}
			\end{minipage}
			\label{fig:example1}
		}
		\subfloat[]{
			\begin{minipage}[t]{0.4\columnwidth}
				\centering
				\includegraphics[width=1.2in,height=1.2in]{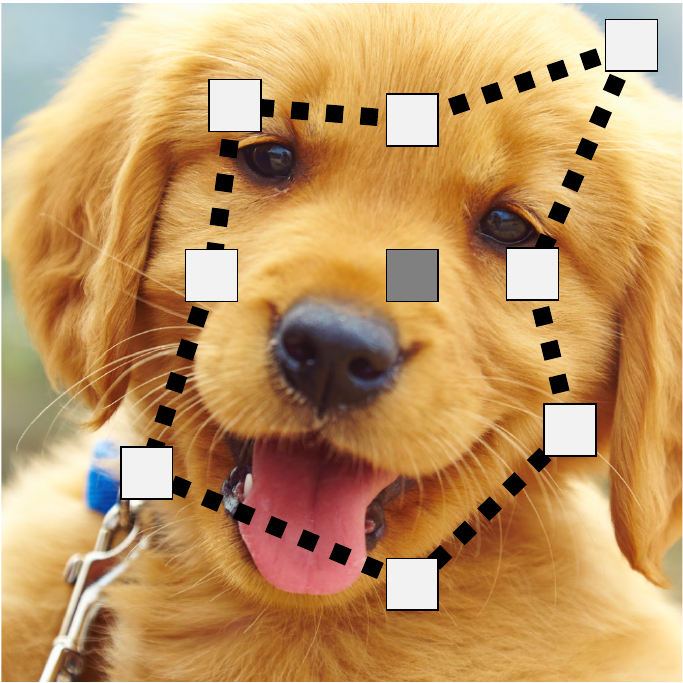}
			\end{minipage}
			\label{fig:example2}
		}
	\end{center}
  \vspace*{-2mm}
	\caption{(a) The fixed receptive field of the dilated convolution, and (b) the dynamic receptive field of the proposed Gaussian dynamic convolution.}\label{fig:example} 
\end{figure}

Researchers have already realized that the multi-scale features and the large receptive field \cite{2015arXiv151107122Y, 2016arXiv161201105Z} are profitable for segmentation tasks. Usually a pyramid architecture is adopted with some dilated convolution or some pooling operation to obtain the multi-scale features with a large receptive field. Figure ~\ref{fig:example1} illustrates the receptive field of the dilated convolution using a toy example. Dilated convolution can expand the receptive field but it can only supply the same scale features. By contrast, our GDC is capable of overcoming the limitation of the dilation factors. The receptive field of the GDC is illustrated in Fig.~\ref{fig:example2}, where the convolution kernel dynamically selects convolutional positions, and the weight vectors are scattered to different positions to extract features in diverse scales. Consequently, the dynamic convolution kernel can not only extend the receptive field but also produce richer feature maps with variable receptive fields.

More specifically, the GDC stochastically forms the convolution kernel of various spatial scales according to some Gaussian distribution offsets. Owing to this randomness, the GDC can supply more diverse feature maps to the following networks. We first employ the GDC to build a lightweight segmentation network for the single image segmentation. Furthermore, we also introduce the GDC to build a Gaussian dynamic pyramid pooling (GDPP) module for the traditional semantic segmentation task, in order to show its generality. The experiments conducted demonstrate that the proposed GDC achieves competitive object semantic segmentation results on the Pascal-Context \cite{Mottaghi:2014ie}, Pascal-VOC 2012 \cite{Everingham:2010dq}, and Cityscapes datasets \cite{Cordts2016Cityscapes}. We also conduct some explanatory experiments to discuss why the proposed GDC works. 

In summary, the main contributions of this paper are:

\begin{itemize}
	\item We propose a novel GDC with a dynamic receptive field to aggregate richer features in various scales.
	\item Our GDC can be easily and efficiently inserted into various convolutional modules for better segmentation performance.
	\item Our experiments indicate that the GDC achieves the state-of-the-art performance in the single image segmentation and semantic segmentation tasks on the three datasets, Pascal-Context, Pascal-VOC 2012, and Cityscapes.
	\item We explain the mechanism and properties of the proposed GDC by additional experiments.
\end{itemize}

The rest of this paper is organized as follows. We briefly introduce the related work in Section~\ref{sec:related}. Our GDC is proposed in Section~\ref{sec:gdc}. Section~\ref{sec:single} introduces the two typical segmentation networks with our GDC. The experiments are conducted in Section~\ref{sec:exp}. In Section~\ref{sec:dis}, we analyze the mechanism of of this novel GDC and discuss why it works. Finally, our conclusions are offered in Section~\ref{S7}.

\section{Related Work}\label{sec:related} 

Semantic segmentation is one of the fundamental tasks in computer vision and it benefits a variety of applications, ranging from biometric identification to object recognition \cite{QWang2019}. The essence of semantic segmentation is to identify distinctive features for different categories. Researchers in the early years tried to extract the handcrafted features, such as the color feature or the geometric feature, to discriminate the labels of all the pixels. Thanks to the resurgence of deep convolutional neural networks (DCNNs), segmentation technology has made great progress in the past few years. Driven by powerful DCNNs \cite{Krizhevsky:2012wl, 2014arXiv1409.1556S, Szegedy:2015gt,  He:2016ib}, the deep segmentation networks, such as FCN \cite{2014arXiv1411.4038L}, SegNet \cite{Badrinarayanan:2017is}, PSPNet \cite{2016arXiv161201105Z}, and DeepLabs \cite{Chen:2018bf, Chen:2018wj}, achieve the state-of-the-art performance of the semantic segmentation task on different benchmark datasets. In particular, FCN \cite{2014arXiv1411.4038L} adopts a fully convolutional network and is optimized by end-to-end training. SegNet \cite{Badrinarayanan:2017is} shares the same idea to design an encoder-decoder architecture and uses some skip connections to utilize the low-level features. For more fine tasks, skip connections are not enough to help accurately locating indistinct boundaries. Zhou \textit{et al.} \cite{8741187} proposed a novel high-resolution multi-scale encoder-decoder network, in which multi-scale dense connections are introduced for the encoder-decoder structure to exploit comprehensive semantic information. SSAP \cite{9056852} uses the pixel-pair affinity pyramid, combines the affinity of pixel pairs and semantic segmentation at different scales, and improves the predictions of instances level by level starting from the deepest layer. Wang \textit{et al.} \cite{8598722} focus on multi-level features to object segmentation. Their conditional Boltzmann machine is suitable to map multi-level convolutional features of object parts onto the global shape of object. Furthermore, context representations have been widely used to profit semantic image segmentation. SCN \cite{8584494} novelly uses the local structural feature maps to compute the context representations in top-down switchable information propagation. The context representations are combined with the convolutional features to form the intermediate feature maps, which are used for the final semantic segmentation. Ji \textit{et al.} \cite{8943115} proposed locality-preserving CNN, which uses relationship between similar samples to auxiliary segmentation. Experiments show that locality-preserving is more suitable for small sample segmentation process. Lin \textit{et al.} \cite{9094232} proposed cross domain complex learning, which effectively utilizes the segmentation labels of synthetic images and variation of real images through an auxiliary task. This method has been proved to be able to transfer context information knowledge from domain to domain in some specific tasks

In contrast to the semantic segmentation, single image segmentation does not rely on a training dataset. Instead, the segmenting method needs to be optimized on each image independently. This task is vital in interactive segmentation \cite{Boykov:1999ug, Tang:2013jo ,9069887} and weakly supervised segmentation \cite{Zongpu2018}. In recent years, many researches \cite{Papandreou:2015fs, SUN2020105824, Ahn:2018vs, Zhou2018, 2018arXiv180306503L, Khoreva2016,2020arXiv191108039Z} focus on weakly supervised segmentation to conquer the problem of scarcity of labeled data. One of the mainstream strategies for this task is to transform the weak labels, such as points \cite{2016arXiv150602106B, 2018arXiv171109081M}, scribble \cite{2016arXiv160405144L,2018arXiv180401346T}, bounding box \cite{Khoreva2016, 7410548} or image-level label \cite{Wei2015}, to a coarse segmentation ground truth for the fully supervised segmentation networks. Hong \textit{et al.} \cite{8103348} provide a comprehensive overview of weakly supervised approaches for semantic segmentation. Specifically, they point out the limitations of various supervision level methods and discuss the directions worthy of effort to improve performance.

For the weakly supervised segmentation based on image-level labels, STC \cite{Wei2015} adopts a simple-to-complex framework and proposes a method for obtaining refined segmentation results with three progressively powerful DCNNs. Redondo \textit{et al.} \cite{8651353} propose a hide-and-seek strategy which consists of two class activation mapping (CAM) modules so as to recover the activation masks covering the full object extents by randomly hiding patches in a training image, forcing the second CAM network to seek other relevant parts. Chen \textit{et al.} \cite{chen2020r} utilize a self-supervised scheme without any ground truth to promote the saliency detection and image segmentation results. Meng \textit{et al.} \cite{8941066} Proposed a new segmentation strategy, which first segmented the foreground by class-level, and then fused all the foreground information to get the final result. The weaklier supervised semantic segmentation \cite{8782155} imposes even more challenges, as it uses only one image-level annotation per category to achieve a desired semantic segmentation performance. Furthermore, for each category, one sample has image level annotation, while only the number of object categories contained in each image is provided for other samples. Mixed-use of image-level labels and bounding box labels can further improve performance \cite{8438912}.

For scribble supervision, ScribbleSup \cite{2016arXiv160405144L} uses some scribbles to generate the segmentation results for an image. GraphNet \cite{Pu:2018hy} combines the deep network with the graph structure. These methods use the scribble annotations to generate the pseudo annotations by the graph convolution. However, ScribbleSup generates segmentation proposals from scribbles and uses these proposals to alternatively train an FCN, which can easily be trapped in local minimums. To solve this problem, Tang \textit{et al.} \cite{2018arXiv180401346T} abandon the alternating training method and train a FCN via a joint loss function with two terms: the partial cross-entropy loss for scribbles only and the relaxed normalized-cut regularizer that implicitly propagates the true labels to unknown pixels during training. Shen \textit{et al.} \cite{8953706} jointly train the weakly supervised object detection and segmentation tasks to complement each other's learning. Such a cross task enforcement helps both the tasks to leap out of their respective local minimums. The work \cite{WangBin19} presents a boundary perception guidance (BPG) approach that only leverages scribbles. It consists of two basic components which are prediction refinement and boundary regression to make better segmentation progressively. Wang \textit{et al.} \cite{8270673} incorporate convolutional neural networks (CNNs) into a bounding box and scribble-based binary segmentation pipeline to resolve the problem of interactive 2D and 3D medical image segmentation. Ji \textit{et al.} \cite{Zhanghexuan2019Scribble} present a scribble-based hierarchical weakly supervised learning pipeline for medical image structure segmentation which integrates graph-based method with only whole tumor/normal brain scribbles and the global labels. This work was the first to realize such a weak supervision level in the field of compression structure segmentation.

There is a contradiction in deep networks for segmentation. Receptive field plays a critical role in the segmentation task. Conventional deep convolution networks expand the receptive field through stacking plenty of convolution layers and pooling layers. This strategy exposes a defect. There is a huge resolution disparity between the feature maps and the input images. Many variations of the normal convolution attempt to resolve this paradox. Dilated convolution \cite{2015arXiv151107122Y} inserts some gaps into the convolution kernel to support an exponential expansion of the receptive field. Deformable convolution \cite{2017arXiv170306211D}, which is commonly used for detection, adopts a side network to learn an offset for each weight vector in the convolution kernel. Both these two approaches can effectively expand the receptive field without losing the resolution of the feature maps. Another essential factor for segmentation is the multi-scale features. Researches such as PSPNet \cite{2016arXiv161201105Z} and DeepLabs \cite{Chen:2018wj} endeavor to extract the multi-scale features through a pyramid architecture. They use different dilated factors or different pooling strides in different layers of the pyramid to capture multi-scale features.

Almost all the existing methods of image segmentation discussed in this section use normal or dilated convolution. Different from dilated convolution or deformable convolution, our Gaussian dynamic convolution is simpler and more effective in many cases. Specifically, our GDC dynamically changes the offsets of the weight vectors in the convolution kernel. Owing to its flexible dynamic range of receptive fields, our GDC can generate rich features. In this paper, we use our GDC to build a lightweight CNN to accomplish the labels transformation. Our lightweight segmentation network only requires one image with some seeds such as the scribbles, and it can be optimized in a flash. Since our GDC introduces dynamics and random factors, it outperforms the existing convolution methods for the one-image segmentation with some seeds such as the scribbles. Moreover, our GDPP built on the proposed GDC offers a competitive means to the existing state-of-the-arts method for the semantic segmentation.

\section{Gaussian Dynamic Convolution}\label{sec:gdc} 

Without loss of generality, we use the $3\times 3$ convolution kernel as an example to illustrate the proposed GDC. 

To highlight the difference of our GDC with other types of convolution, we start from the normal convolution kernel, where a regular $3\times3$ grid slides over the feature maps $F^l$ at layer $l$ to sample 9 feature vectors. A normal convolution kernel summarizes the sampled vectors weighted by the corresponding weights in the convolution kernel. Let $c=<i, j>$ be the coordinate of the center feature vector in one convolution operation. Then, the coordinate $\overline{c}$ of the other eight sampled feature vectors can be calculated with a direction basis $e$ and an offset value $\Delta$ according to
\begin{align}\label{conv_offset}  
  \overline{c}_i =& c + \Delta_i \odot e_i , \nonumber \\ 
	e_i \in & {\cal E} = \{<-1,-1>,<-1,0>,\cdots , <1,1>\} ,
\end{align}
where the direction basis $e$ denotes the basic sample direction, the offset value $\Delta$ is always $<1, 1>$ in a normal $3\times3$ convolution kernel, and $\odot$ denotes the element-wise product operator. The element-wise product may be replaced by the Hadamard product. For example, the coordinate of the sampled feature vector at the left-top corner can be calculated by:
\begin{equation}\label{eq2}
	<i, j>+<1, 1>\odot<-1, -1>=<i-1,j-1> .
\end{equation}

\begin{figure}[h]
\vspace*{-3mm}
	\begin{center}
		\includegraphics[width=\columnwidth]{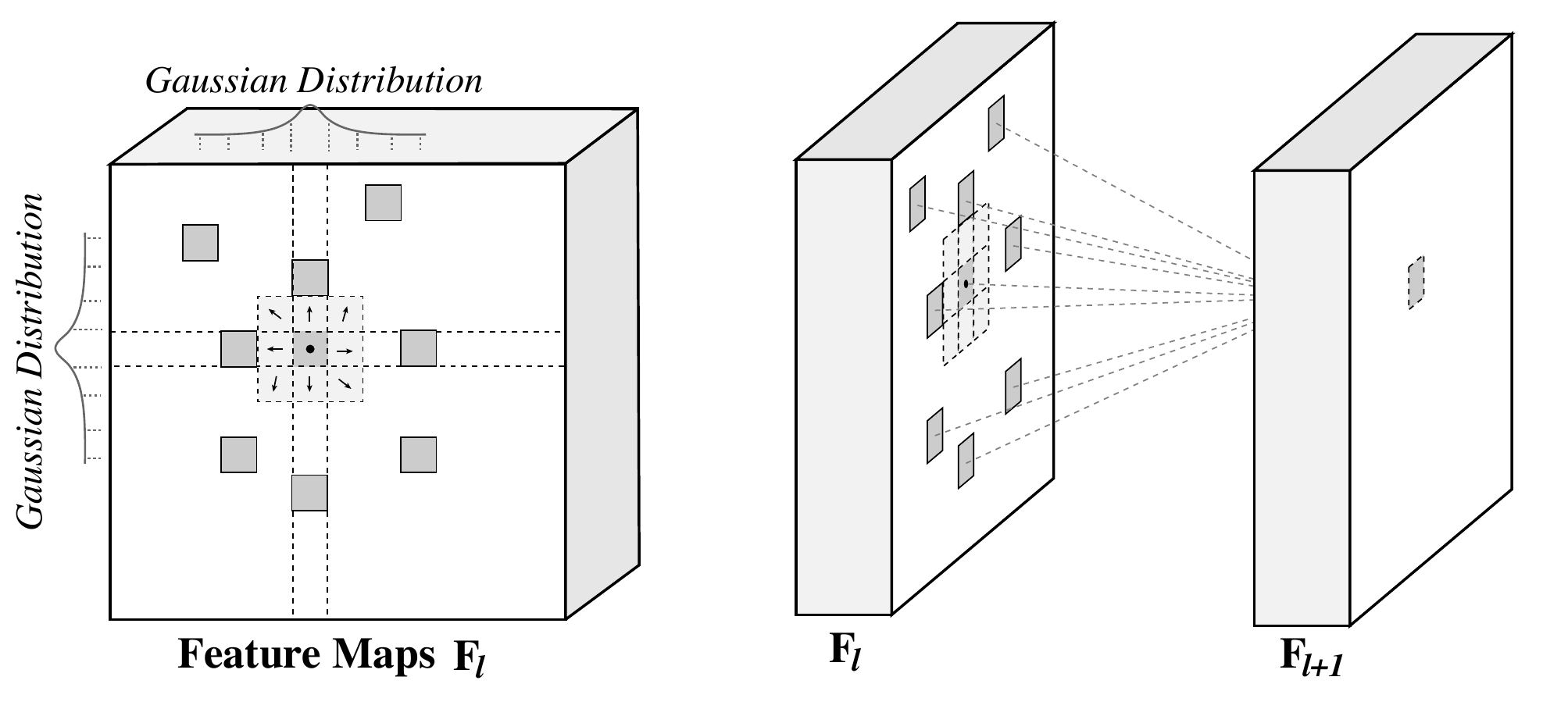}
	\end{center}
\vspace*{-6mm}
	\caption{Illustration of the Gaussian dynamic convolution with kernel size 3.}
	\label{fig:fig1} 
\vspace*{1mm}
\end{figure}

For the dilated convolution \cite{2015arXiv151107122Y}, we can also calculate the coordinates of the sampled feature vectors by Eq.\,(\ref{conv_offset}). The dilated factor is equal to the offset value $\Delta$.

\begin{figure*}[tp]
\vspace*{-1mm}
	\begin{center}
		\includegraphics[width=\textwidth]{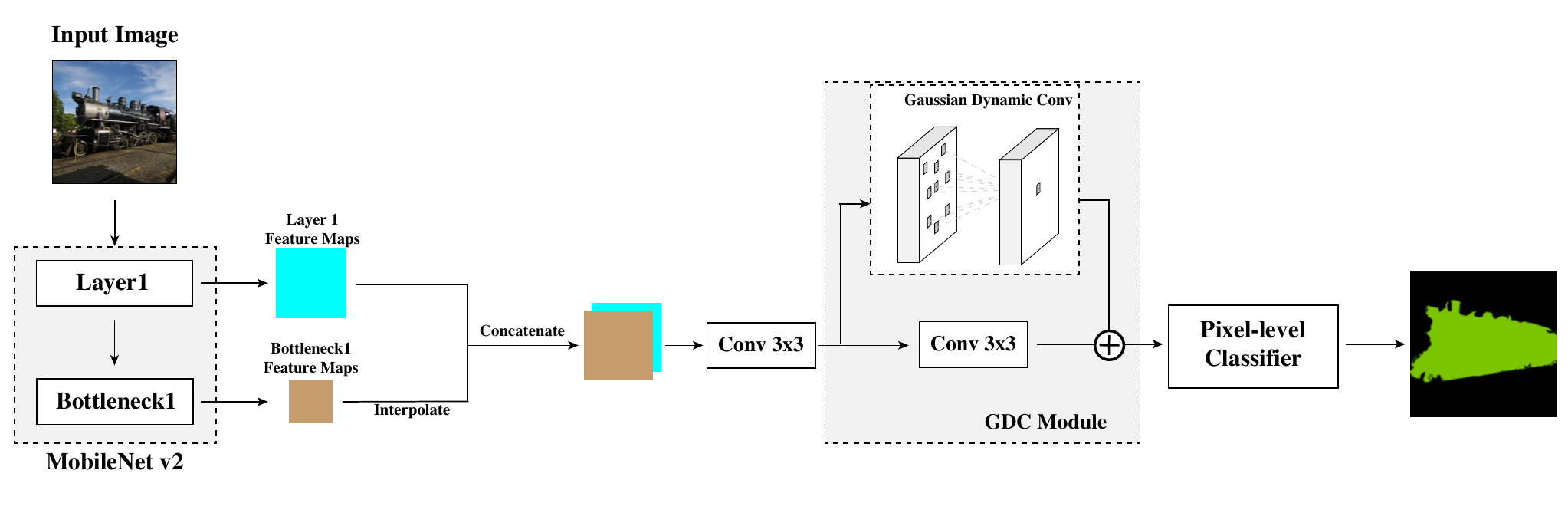}
	\end{center}
\vspace*{-8mm}
	\caption{Pipeline of the lightweight segmentation network with the Gaussian dynamic convolution. We adopt the first layer and the first convolution linear bottleneck of the MobileNet v2 to extract two groups of feature maps.}
	\label{fig:pipeline} 
\vspace*{-2mm}
\end{figure*}

The proposed GDC is illustrated in Fig.~\ref{fig:fig1}. For each convolution position, we fix the center weight of the convolution kernel. Then we scatter the other convolution weight vectors by randomly changing the offset value $\Delta$. More specifically, we sample the offset values from a two-dimensional half Gaussian distribution with standard deviation $\Sigma$, denoted by $HalfGaussian_2(0,\Sigma)$. 
The one-dimensional half Gaussian distribution is expressed as:
\begin{equation}\label{eq3}
	HalfGaussian(0, \Sigma)=\frac{\sqrt{2}}{\Sigma \sqrt{\pi}}\exp\left(-\frac{x^2}{2\Sigma^2}\right), x \geq 0.
\end{equation}
Suppose that $c$ is the coordinate of the center convolution position. The coordinates $\tilde{c}$ of the other convolution positions in the GDC can also be calculated by Eq.\,(\ref{conv_offset}). But instead of using a fixed offset value $\Delta =<1, 1>$, $\Delta$ in our GDC obeys $HalfGaussian_2(0, \Sigma)$. Different $\Delta$ values will produce different feature maps. The summation operation of the sampled feature vectors in the GDC is the same as in the normal convolution.

\section{Gaussian Dynamic Convolution based Image Segmentation Networks}\label{sec:single} 

\subsection{Single-Image Segmentation}\label{S4.1}

Single image segmentation is quite different from the traditional semantic segmentation. Usually, it needs to optimize the algorithm and generates the segmentation result all based on only one image. For the single image segmentation task, DCNNs are inapplicable. This is because conventional deep segmentation networks are typically composed of plenty of convolutional layers to obtain more information. These DCNNs need to be trained with a large set of data. One image is insufficient to support the training of a deep network, and the optimization time of a deep network is too high. Ideally, a lightweight network is preferred, which can be optimized extremely fast and can generate satisfactory segmentation result. In this subsection, we address the challenges imposed by the single image segmentation task, and design a lightweight segmentation network with the novel GDC proposed in the previous section. This lightweight network based on the GDC can be optimized in a flash and it is capable of generating satisfactory segmentation result.

\subsubsection{Network Architecture}\label{S4.1.1}

As shown in Fig.~\ref{fig:pipeline}, we use the first layer and the first convolutional linear bottleneck of the MobileNet v2 \cite{2018arXiv180104381S} as our feature extractor. These two network components generate two groups of feature maps with only 16 and 24 channels, respectively. We resize these two groups of feature maps to the half size of the input image by the bilinear interpolation. Then, the feature maps are concatenated and sent into a $1\times1$ convolutional layer to fuse the channel information. After that, we send the fused feature maps to the GDC module.

The GDC Module consists of two branches. The first one is a normal $3\times3$ convolution layer which is used to fuse the local features. The second branch is our GDC. We use this branch to gather features in various scales.

Finally, we send the feature maps generated by the GDC module to a pixel-level classifier to estimate the final segmentation result. The pixel-level classifier is composed of a $3\times3$ convolution layer followed by a ReLU activate function, and a $1\times1$ convolution layer with a softmax layer.

\subsubsection{Training}\label{S4.1.2}

We initialize and optimize our lightweight segmentation network for each image, instead of using plenty of training data. In our experiment, the training ground truth is generated by some weakly semantic cues such as the scribbles \cite{2016arXiv160405144L}. This segmentation network is extremely fast as it consists of very few convolution layers. The detailed training of this lightweight GDC segmentation network is given in Section~\ref{sec:exp}.

\begin{figure}[bp]
	\vspace*{-5mm}
	\begin{center}
		\includegraphics[width=0.76\columnwidth]{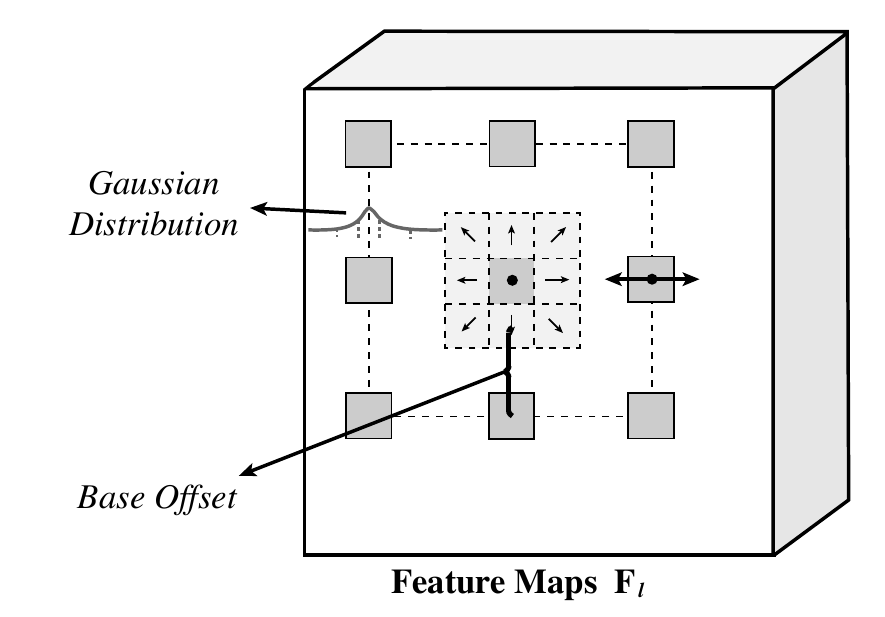}
	\end{center}
	\vspace*{-6mm}
	\caption{Illustration of the variant Gaussian dynamic convolution with kernel size 3.}
	\label{fig:semidyn} 
\end{figure}

\subsection{Semantic Segmentation}\label{S4.2}

As discussed previously, multi-scale features are essential for achieving good performance in semantic segmentation tasks. Traditionally a pyramid architecture is adopted with dilated convolution to obtain the multi-scale features with a large receptive field. In order to show the potential and generality of the proposed GDC , we further design a variant GDC as shown in Fig.~\ref{fig:semidyn} suitable for the application to semantic segmentation. Specifically, we scatter the weight vectors of the convolution kernel by a same offset value $\Delta$. The offset values $\Delta$ are sampled from a two-dimensional half Gaussian distribution with standard deviation $\Sigma$ as shown in Fig.~\ref{fig:semidyn}. We set a constant base offset and calculate the sampled position by introduce a hyperparameter $\Delta_{base}$ in the variant GDC. The sampled position can be calculated by:
\begin{equation}\label{conv_offset_semi} 
	\overline{c}_i = c + (\Delta_{base} + \Delta) \odot e_i ,
\end{equation}
where $c$ is the coordinate of the center convolution position.

It is clear that this variant GDC degenerates into the dilated convolution when the dynamic offset $\Delta =0$. The constant base offset or hyperparameter $\Delta_{base}$ ensures that our dynamic convolution can expand the receptive field as the dilated convolution does. Moreover, the dynamic offset $\Delta$ brings richer features for the following network components. We adopt this variant GDC to build a GDPP for implementing the semantic segmentation network. 

\begin{figure}[hp]
\vspace*{-2mm}
	\begin{center}
		\includegraphics[width=0.8\columnwidth]{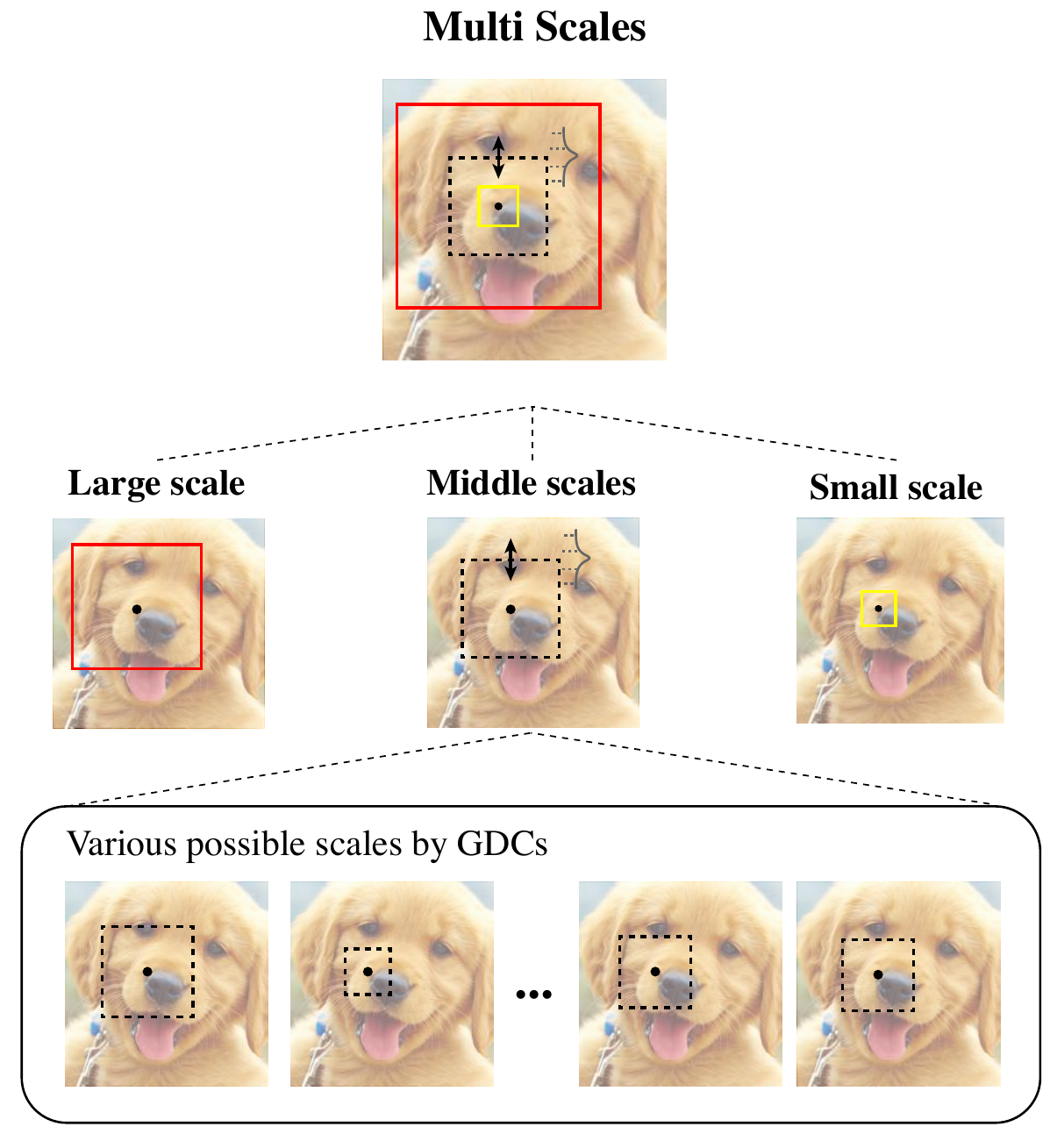}
	\end{center}
\vspace*{-5mm}
	\caption{Illustration of the Gaussian dynamic pyramid pooling, where the largest and smallest scales are fixed, while the middle scales are dynamically sampled by the Gaussian distribution.}
	\label{fig:gdpp_illus}
\end{figure} 

Multi-scale feature fusing modules for semantic segmentation usually use several dilated convolution layers with different factors to gather multi-scale information. In Fig.~\ref{fig:gdpp_illus}, the small yellow square denotes the dilated convolution layer with a small factor which can be used to gather the small scale information. On the other hand, the dilated convolution denoted by the large red square is used to obtain large scale information. To fuse the middle-scale information, other modules, such as ASPP \cite{Chen:2018wj}, use one or more dilated convolutions with different fixed middle factors. Our GDPP module is different in that we use the GDCs to gather the information to implement various middle scales, as shown in the bottom part of Fig.~\ref{fig:gdpp_illus}. In other words, in our GDPP module for semantic segmentation depicted in Fig.~\ref{fig:gdpp_illus}, the largest and smallest scales are fixed to limit the range of GDCs, while the middle scales are dynamically sampled by the Gaussian distribution. Therefore, the proposed GDPP module can produce richer and more vivid feature maps during the training phase.

\section{Experiments}\label{sec:exp} 

The experiments (both training and testing) were conducted on a PC with a GTX 2080ti GPU. The implementation code will be published on {https://github.com/ouc-ocean-group/}.

\begin{table*}[tp]
	\caption{Performance comparison of the single image segmentation on two different datasets of Pascal Context and Pascal-VOC 2012.}
	\label{tab:single-performance} 
	\vspace*{-4mm}
	\begin{center}
		\begin{tabular}{p{0.3\columnwidth}|p{0.22\columnwidth}p{0.25\columnwidth}|p{0.22\columnwidth}p{0.25\columnwidth}}
			\toprule
			& \multicolumn{2}{c}{\textbf{Pascal Context}} & \multicolumn{2}{|c}{\textbf{Pascal-VOC 2012 (val)}}\\
			\cmidrule{2-5}
			\textbf{Methods} & \textbf{Overall Acc.} & \textbf{mIoU (\%)} & \textbf{Overall Acc.} & \textbf{mIoU (\%)} \\
			\midrule
			Normal 3x3 conv & 0.7900 & 53.41 & 0.8887 & 68.94 \\
			\midrule
			Dilated conv (6) & 0.8327 & 60.01 & 0.9047 & 71.95 \\
			Dilated conv (16) & 0.8324 & 59.86 & 0.9034 & 71.80 \\
			Dilated conv (24) & 0.8308 & 59.57 & 0.9029 & 71.78 \\
			\midrule
			Deformable conv & 0.8003 & 54.67 & 0.8883 & 68.69 \\
			\midrule
			\textbf{GDC ($\Sigma=0.2$)} & \textbf{0.8646} & \textbf{65.12}\ $\blacktriangle 11.71$ & \textbf{0.9092} & \textbf{74.06}\ $\blacktriangle 5.12$ \\
			\bottomrule
		\end{tabular}
	\end{center}
	\vspace*{-5mm}
\end{table*}

\begin{figure}[tp]
	\begin{center}
		\includegraphics[width=\columnwidth]{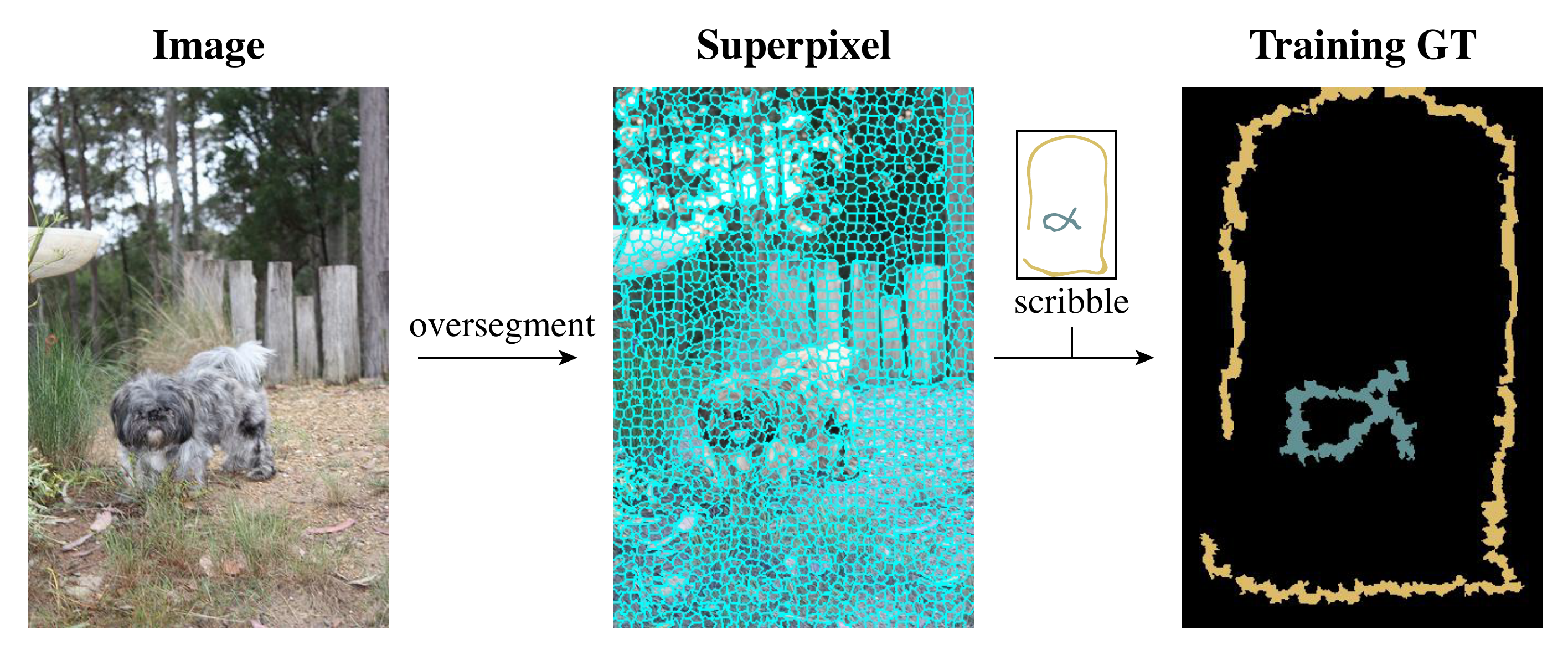}
	\end{center}
	\vspace*{-4mm}
	\caption{Illustration of the training ground truth expansion.}
	\label{fig:training-gt} 
	\vspace*{-4mm}
\end{figure}

\subsection{Single Image Segmentation}\label{S5.1}

\subsubsection{Implementation}\label{S5.1.1}

Single image segmentation plays a vital role in many other tasks including weakly supervised segmentation label generation. The sizes of the images in single image segmentation are variable. Therefore, we adopt an adaptive Gaussian dynamic offset of $s\times\Delta$, where $s$ is the length of the shortest side of the image, and $\Delta$ is the offset value sampled from a half Gaussian distribution. The lightweight segmentation network is initialized for each image. The MobileNet v2 head is pre-trained on the ImageNet dataset \cite{Deng:2009dl}. We fix all the parameters of it and optimize the rest components for 50 steps using the ground truth generated by the scribble. We use the scribble annotations provided by \cite{2016arXiv160405144L}. In order to expand the labeled areas of these scribbles, as illustrated in Fig.~\ref{fig:training-gt}, we over-segment the input image into superpixels by SLIC \cite{Achanta:2012eg}. The class labels of these superpixels depend on the scribbles. For a superpixel $SP_j$, if there is a scribble $SC_i$ with class $c_i$ overlap $SP_j$, we mark all the pixels in $SP_j$ with label $c_i$. The loss function we employed is the weighted cross-entropy loss:
\begin{equation}\label{eq5}
	CELoss_i = - \sum_{i=1}^N w_{c_i} y_{c_i} \log(p_i) ,
\end{equation}
where $N$ is the number of categories, $p_i$ denotes the prediction probability of class $c_i$, and $y_{c_i}=1$ indicates the ground truth of this prediction is $c_i$, while we calculate the weight for each category by: 
\begin{equation}\label{eq6}
	w_{c_i} = \frac{n_{c_i}}{N_{all}} ,
\end{equation}
in which $n_{c_i}$ is the number of the pixels of category $c_i$ and $N_{all}$ is the number of all the labeled pixels. We employ SGD \cite{Bottou2010} as our network optimizer with the learning rate set to 0.01. When the optimization is completed, the GDC samples extra 50 offset values to generate 50 different segmentation results. We average these 50 results as the final segmentation result. 

\subsubsection{Performance}\label{S5.1.2} 

\emph{a)~The advantage of GDC}: We first evaluate the performance of our GDC based lightweight network on the Pascal-Context dataset \cite{Mottaghi:2014ie} which involves 59 categories of objects and stuff. The accuracy is evaluated by the overall accuracy and the mean Intersection-over-Union (mIoU) score. In order to demonstrate the advantage of the proposed GDC over the other convolutions, we use 3 other different convolution kernels, namely, normal convolution, dilated convolution \cite{2015arXiv151107122Y} and deformable convolution \cite{2017arXiv170306211D}, to replace the GDC layer in our lightweight network to produce the three alternative lightweight networks for comparison. Following the symbol conventions of the previous sections, the output of deformable convolution is calculated by
\begin{equation}\label{eq7}
	output^c=\sum_{e_i\in {\cal E}} w^{e_i}\cdot x^{c+(\Delta_{base}+\Delta_i)\odot e_i},
\end{equation}
where $c$ is the coordinate of the center convolution position, $w^{e_i}$ is the weight of the convolution kernel, and $x^{c+(\Delta_{base}+\Delta_i)\odot e_i}$ is the input feature vector at the coordinate $c+(\Delta_{base}+\Delta_i)\odot e_i$. The convolution position is mainly controlled by a bias $\Delta_i$ and this bias is obtained by a trainable neural network. When $\Delta_i=0$, the deformable convolution degenerates into the dilated convolution.

\begin{figure}[bp]
\vspace*{-6mm}
	\begin{center}
		\captionsetup[subfigure]{labelformat=empty}
		\subfloat[ w/o CRF]{
			\begin{minipage}[b]{0.30\columnwidth}
				\centering
				\includegraphics[width=\columnwidth]{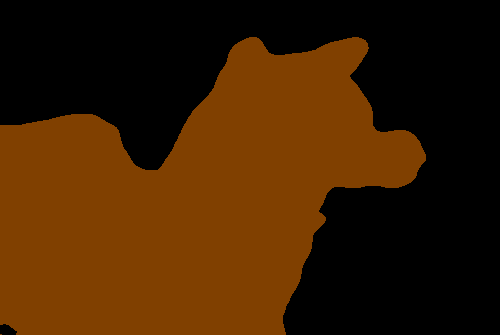} \\
				\includegraphics[width=\columnwidth, height=5em]{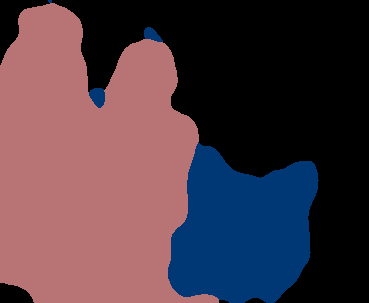} \\
				\includegraphics[width=\columnwidth, height=5em]{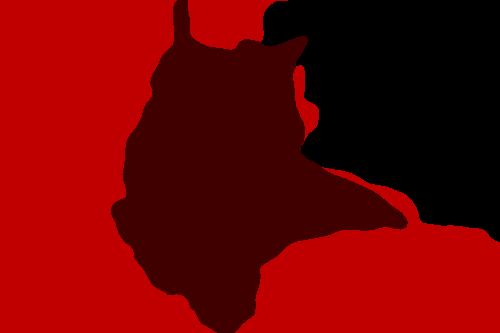}
			\end{minipage}
		}
		\subfloat[  w/ CRF]{
			\begin{minipage}[b]{0.30\columnwidth}
				\centering
				\includegraphics[width=\columnwidth]{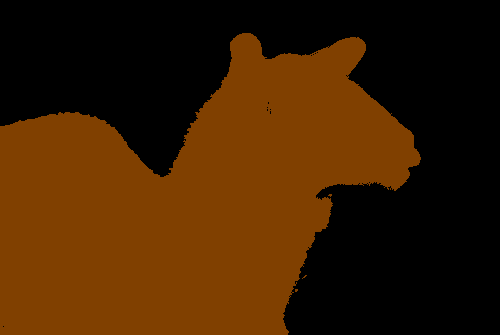} \\
				\includegraphics[width=\columnwidth, height=5em]{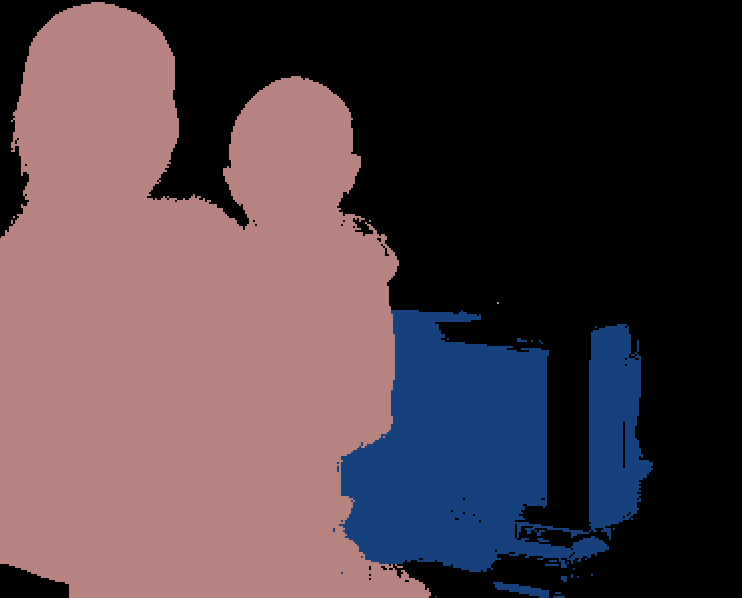} \\
				\includegraphics[width=\columnwidth, height=5em]{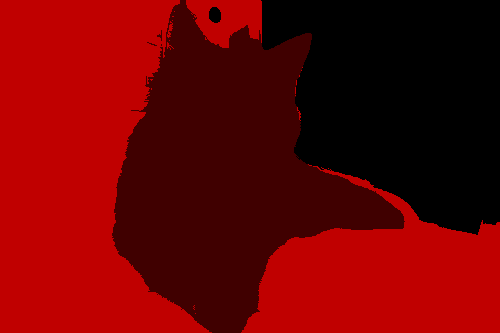}
			\end{minipage}
		}
		\subfloat[  GT (real)]{
			\begin{minipage}[b]{0.30\columnwidth}
				\centering
				\includegraphics[width=\columnwidth]{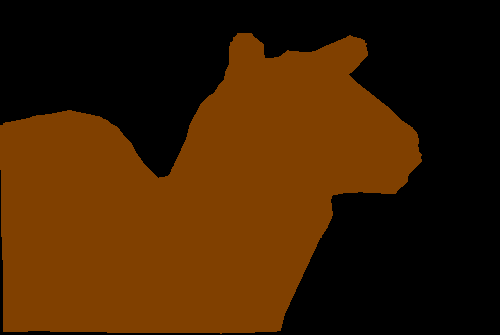} \\
				\includegraphics[width=\columnwidth, height=5em]{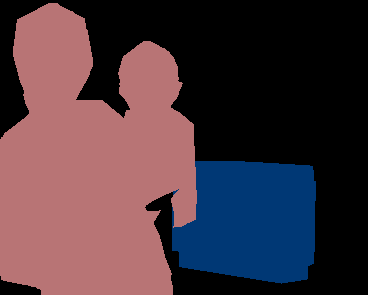} \\
				\includegraphics[width=\columnwidth, height=5em]{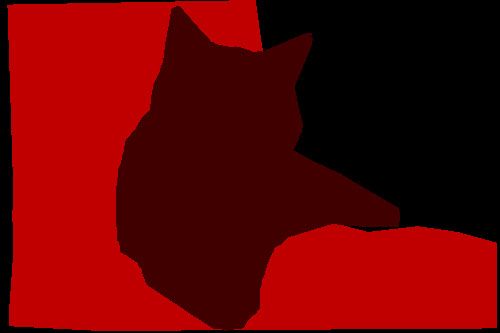}
			\end{minipage}
		}
	\end{center}
	\vspace*{-3mm}
	\caption{Visualization of the ground truth on Pascal-VOC 2012.}
	\label{fig:voc-gt} 
	\vspace*{-1mm}
\end{figure}

As shown in the left part of Table \ref{tab:single-performance}, we use a normal $3\times3$ convolution layer as the baseline. For the Pascal-Context dataset, the normal convolution baseline achieves 53.41\% mIoU. The dilated convolution with $\text{factor}=6$ and deformable convolution increase the mIoU to 60.01\% and 54.67\%, respectively, because they both can expand the receptive field. Notably, the deformable convolution can hardly boost the performance. The possible reason is that the low level feature maps extracted by the one image training based lightweight network is insufficient for learning a reliable offset. By contrast, our GDC achieves the highest mIoU of 65.12\%, which is 11.71\% higher than the baseline. We will discuss the reasons for this performance improvement later.

We also evaluate our method on the validation set of Pascal-VOC 2012 \cite{Everingham:2010dq} which contains 1,449 images with 21 categories. We report the results in the right part of  Table \ref{tab:single-performance}. The normal convolution baseline attains the mIoU of 68.94\%. The dilated convolution with dilated $\text{factor}=6$ and deformable convolution achieve the mIoU values of 71.95\% and 68.69\%, respectively. Our GDC by comparison increases the mIoU to 74.06\%, which is the highest on this dataset among the four lightweight segmentation networks.

\begin{table}[h]
	\vspace*{-3mm}
	\caption{Performance comparison of the segmentation models trained with the pseudo label on Pascal-VOC 2012 validation set. \dag  means refining by CRFs.}
	\label{tab:weakvoc}
	\vspace*{-4mm}
	\begin{center}
		\begin{tabular}{p{0.3\columnwidth}|p{0.3\columnwidth}|p{0.2\columnwidth}{c}}
			\toprule
			\textbf{Methods} & \textbf{Conference} & \textbf{mIoU (\%)} \\
			\midrule
			ScribbleSup\dag & CVPR2016 \cite{2016arXiv160405144L} & 63.1 \\
			RAWKS & CVPR2017 \cite{2018arXiv180401346T} & 61.4 \\
			NormalizedCutLoss & CVPR2018 \cite{Vernaza:2017bl} & 62.4 \\
			GraphNet-Initial & ACMMM2018 \cite{Pu:2018hy} & 63.3 \\
			\midrule
			\textbf{Ours} & - & \textbf{64.9} \\
			\bottomrule
		\end{tabular}
	\end{center}
	\vspace*{-1mm}
\end{table}

\begin{figure}[bp]
\vspace*{-7mm}
	\begin{center}
		\captionsetup[subfigure]{labelformat=empty}
		\subfloat[  Image]{
			\begin{minipage}[b]{0.22\columnwidth}
				\centering
				\includegraphics[width=\columnwidth, height=5em]{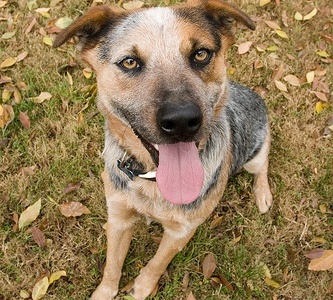} \\
				\includegraphics[width=\columnwidth]{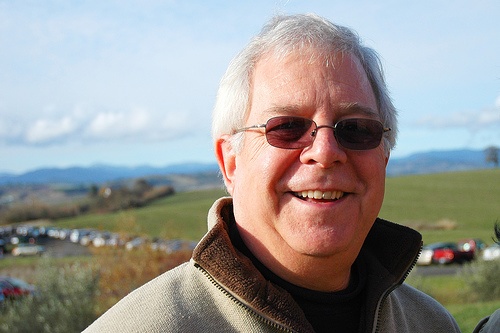} \\
				\includegraphics[width=\columnwidth]{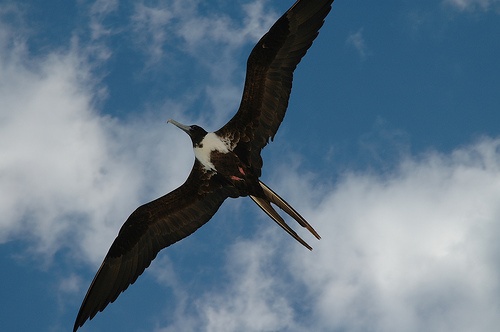} \\
				\includegraphics[width=\columnwidth]{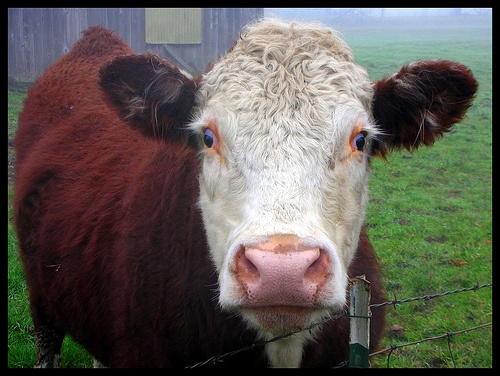}
			\end{minipage}
		}
		\subfloat[  GT]{
			\begin{minipage}[b]{0.22\columnwidth}
				\centering
				\includegraphics[width=\columnwidth, height=5em]{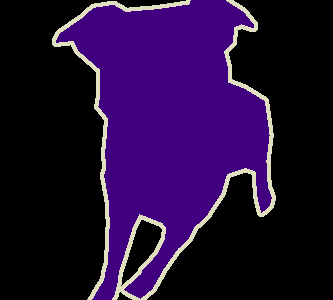} \\
				\includegraphics[width=\columnwidth]{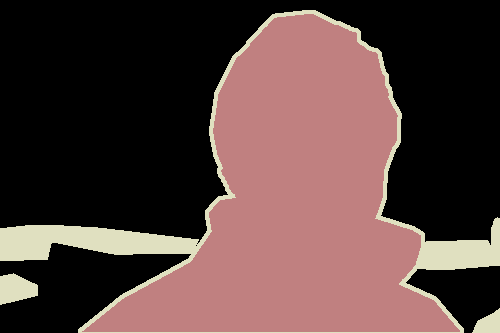} \\
				\includegraphics[width=\columnwidth]{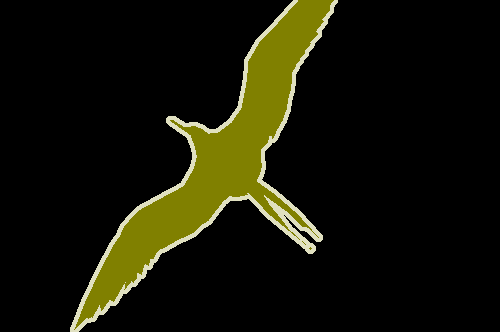} \\
				\includegraphics[width=\columnwidth]{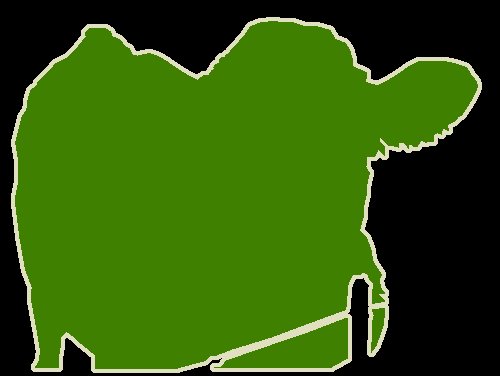}
			\end{minipage}
		}
		\subfloat[ Strong]{
			\begin{minipage}[b]{0.22\columnwidth}
				\centering
				\includegraphics[width=\columnwidth, height=5em]{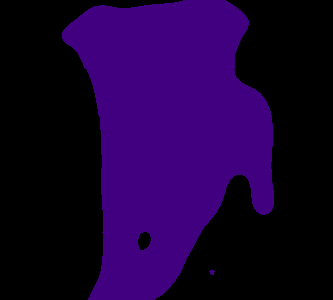} \\
				\includegraphics[width=\columnwidth]{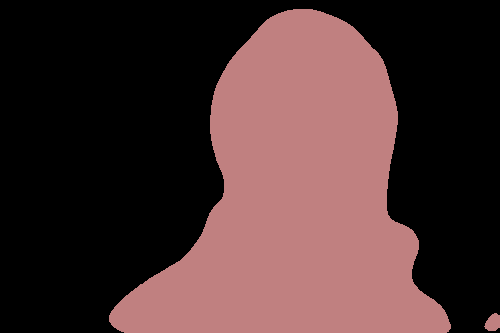} \\
				\includegraphics[width=\columnwidth]{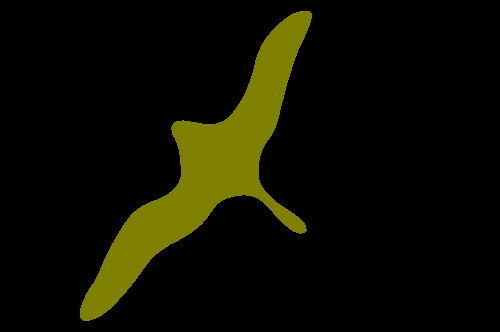} \\
				\includegraphics[width=\columnwidth]{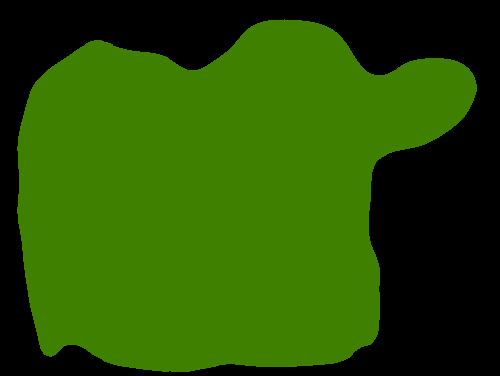}
			\end{minipage}
		}
		\subfloat[ Ours (Weakly)]{
			\begin{minipage}[b]{0.22\columnwidth}
				\centering
				\includegraphics[width=\columnwidth, height=5em]{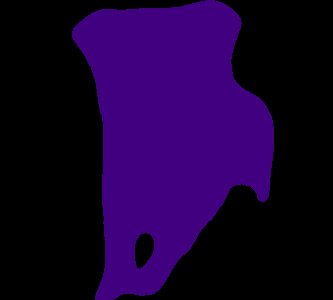} \\
				\includegraphics[width=\columnwidth]{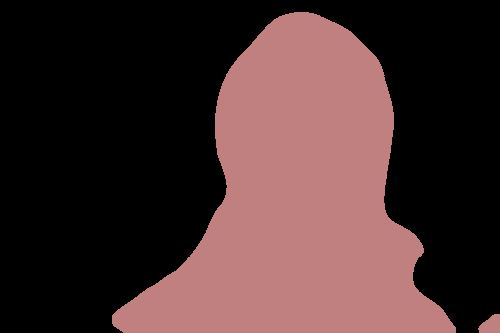} \\
				\includegraphics[width=\columnwidth]{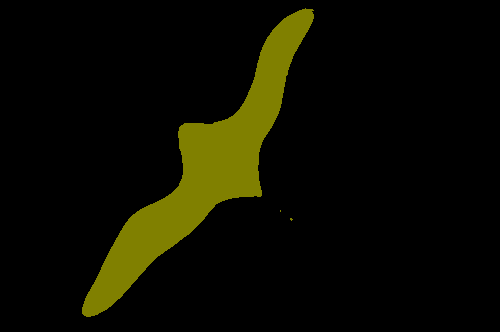} \\
				\includegraphics[width=\columnwidth]{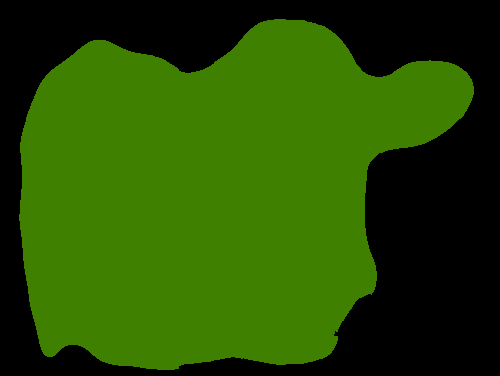}
			\end{minipage}
		}
	\end{center}
	\vspace*{-3mm}
	\caption{Visualization results of ground truth, strong supervision and our weakly supervised segmentation on Pascal-VOC 2012 validation set.}
	\label{fig:voc-weak} 
	\vspace*{-1mm}
\end{figure}

\emph{b)~The quality of the single image segmentation results}: Following GraphNet \cite{Pu:2018hy}, we use the augmented data by Hariharan \textit{et al.} \cite{Hariharan:2011ep} to setup a weakly supervised segmentation experiment to evaluate the quality of our single image segmentation results. We use the single image segmentation results as the pseudo training label to train a DeepLab-v1 \cite{Chen:2018bf} on the training sets of Pascal-VOC 2012 and Pascal Context, respectively. The visualization of the pseudo ground truth and the real ground truth on Pascal-VOC 2012 training set is depicted in Fig.~\ref{fig:voc-gt}. We compare our method with the four state-of-the-art scribble supervised segmentation methods, i.e.,  ScribbleSup \cite{2016arXiv160405144L}, RAWKS \cite{Vernaza:2017bl}, NormalizedCutLoss \cite{2018arXiv180401346T} and GraphNet \cite{Pu:2018hy} on the validation subset. The performance on Pascal-VOC 2012 are reported in Table \ref{tab:weakvoc}. Our method achieves the best mIoU of 64.9\%. Fig.~\ref{fig:voc-weak} shows that the segmentation results generated by our weakly supervised model are comparable to those with strong supervision. This demonstrates the quality of the weakly labels produced by our GDC one image segmentation. Table~\ref{tab:weakcontext} compares the mIoU results of our method with GraphNet-Initial \cite{Pu:2018hy} on Pascal Context validation set. It can be seen that our method also achieves better performance.

\begin{table}[tp]
	\caption{Performance comparison of two segmentation models trained with the pseudo label on Pascal-Context validation set.}
	\label{tab:weakcontext} 
	\vspace*{-3mm}
	\begin{center}
		\begin{tabular}{p{0.3\columnwidth}|p{0.3\columnwidth}|p{0.2\columnwidth}{c}}
			\toprule
			\textbf{Methods} & \textbf{Conference} & \textbf{mIoU (\%)} \\
			\midrule
			GraphNet-Initial & ACMMM2018 \cite{Pu:2018hy} & 33.1 \\
			\textbf{Ours} & - & \textbf{34.1} \\
			\bottomrule
		\end{tabular}
	\end{center}
	\vspace*{-4mm}
\end{table}

\emph{c)~Ablation}: We further investigate the effects of standard deviation $\Sigma$ on the achievable performance of our GDC using Pascal-Context in Table~\ref{tab:sigma-abilation}. The value of $\Sigma$ controls the overall scale of the receptive field. Comparing Table~\ref{tab:sigma-abilation} with Table~\ref{tab:single-performance}, it can be seen that our GDCs with different $\Sigma$ values all achieve better performance than the three existing convolutions. In particular, with $\Sigma=0.2$, the GDC  achieves the best result. As $\Sigma$ decreases or increases, there is a slight decrease in performance. The reason is that a too small $\Sigma$ results in a small sampling area which is inefficient to aggregate large-scale information, while a too large $\Sigma$ may fuse superfluous noise features.

\begin{table}[hp]
	\vspace*{-1mm}
	\caption{The Overall accuracy and mIoU with different $\Sigma$ on Pascal context dataset.}
	\label{tab:sigma-abilation} 
	\vspace*{-4mm}
	\begin{center}
		\begin{tabular}{p{0.2\columnwidth}|p{0.1\columnwidth}p{0.1\columnwidth}p{0.1\columnwidth}}
			\toprule
			$\Sigma$ & 0.1 & 0.2 & 0.3 \\
			\midrule
			Overall Acc. & 0.8591 & \textbf{0.8646} & 0.8602 \\
			mIoU (\%) & 63.97 & \textbf{65.12} & 64.88\\
			\bottomrule
		\end{tabular}
	\end{center}
	\vspace*{-5mm}
\end{table}

\subsection{Semantic Segmentaion}\label{S5.2}

\subsubsection{Implementation}\label{S5.2.1}

We implement the proposed GDPP module for semantic segmentation. First, its performance is evaluated on Pascal-VOC 2012 dataset. The backbone network is a MobileNet v2 pre-trained on ImageNet dataset \cite{Deng:2009dl}. The feature maps generated by the last convolutional layer of the backbone network are sent to the GDPP module. We adopt one GDC layer and two dilated convolution layers in the GDPP module. The two dilated convolutions aggregate the small and very large scale information, respectively, while the GDC layer produces the rich middle-scale information. Following the setup in DeepLabs, the small and large dilated factors are set to 1 and 18, respectively. The base offset of our variant GDC is set to 9, and the Gaussian standard deviation $\Sigma$ is chosen to be 2. Following the depthwise separable convolution in MobileNet, we modify all the convolution layers in the GDPP module to the separable mode to reduce the number of parameters. The tail segmentation network is the same as Deeplabv3 \cite{Chen:2018wj}. The optimizer is SGD with 0.028 learning rate which also follows the setup of the official DeepLabv3. The training epochs are set to 50 and 100, respectively. We use the original Pascal-VOC 2012 training dataset with 1464 images to accomplish the network optimization.

\begin{table}[tp]
	\caption{Semantic segmentation results of four methods on Pascal-VOC 2012 validation set.}
	\label{tab:voc-result} 
	\vspace*{-3mm}
	\begin{center}
		\begin{tabular}{p{0.18\columnwidth}p{0.20\columnwidth}|p{0.19\columnwidth}p{0.14\columnwidth}}
			\toprule
			\textbf{Methods} & \textbf{Setting} & \textbf{mIoU (\%)} & \textbf{Params.} \\
			\midrule
			MobileNetv2 & &  45.81 & - \\
			\midrule
			+ ASPP & 50 epochs & 63.01 & $33.6\times10^5$ \\
			+ ASPP & 100 epochs & 65.03 $\blacktriangle 2.02$ & $33.6\times10^5$ \\
			\midrule
			+ S-ASPP & 50 epochs & 62.29 & $5.9\times10^5$ \\
			+ S-ASPP & 100 epochs & 63.16 $\blacktriangle 0.87$ & $5.9\times10^5$ \\
			\midrule
			$+$ GDPP & 50 epochs & 62.44 & $5.9\times10^5$ \\
			$+$ GDPP & 100 epochs & 64.71 $\blacktriangle 2.27$ & $5.9\times10^5$ \\
			\bottomrule
		\end{tabular}
	\end{center}
	\vspace*{-4mm}
\end{table}

\subsubsection{Performance}\label{S5.2.2}

The semantic segmentation results of the four methods on Pascal-VOC 2012 validation set are reported in Table~\ref{tab:voc-result}. We evaluate the MobileNet v2 backbone with the tail segmentation network and use the result as the baseline. This baseline network does not have any multi-scale information fusing module and can only get an mIoU value of 45.81\%. First, we set the number of training epochs to 50. The original ASPP module \cite{Chen:2018wj} with $33.6\times10^5$ parameters can boost the mIoU to 63.01\%. Our GDPP module achieves an mIoU of 62.44\%, which is slightly lower than the ASPP, but it has only $5.9\times10^5$ parameters. To be fair, we also replace the GDC in the GDPP module by a separable dilated convolution and call the resulting module as S-ASPP. It can be seen from Table~\ref{tab:voc-result} that this S-ASPP module gets an mIoU of 62.29\% which is slightly lower than our GDPP module.

Next, we set the number of training epochs to 100 and train the three modules again. The results obtained are also presented in Table~\ref{tab:voc-result}, where the corresponding increases in mIoUs over the results of 50-epochs training are also provided. It is worth noting that the advantage of the GDPP over the S-ASPP is clearly revealed. Specifically, the GDPP module achieves an mIoU of 64.71\% which is 1.55\% higher than the S-ASPP. We will discuss the reason in the discussion section.

\begin{figure}[h]
\vspace*{-4mm}
	\begin{center}
		\captionsetup[subfigure]{labelformat=empty}
		\subfloat[  Image]{
			\begin{minipage}[b]{0.30\columnwidth}
				\centering
				\includegraphics[width=\columnwidth]{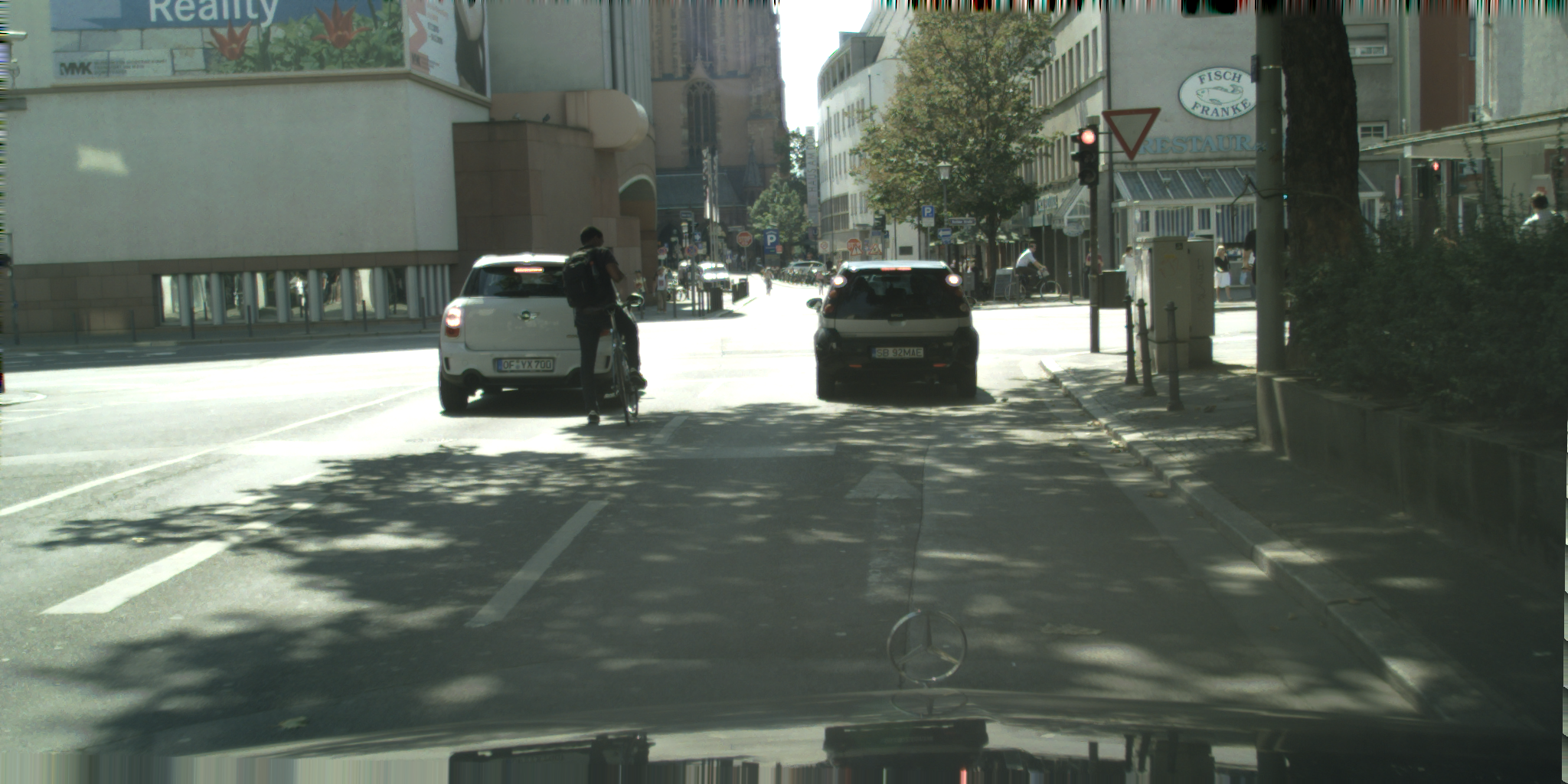} \\
				\includegraphics[width=\columnwidth]{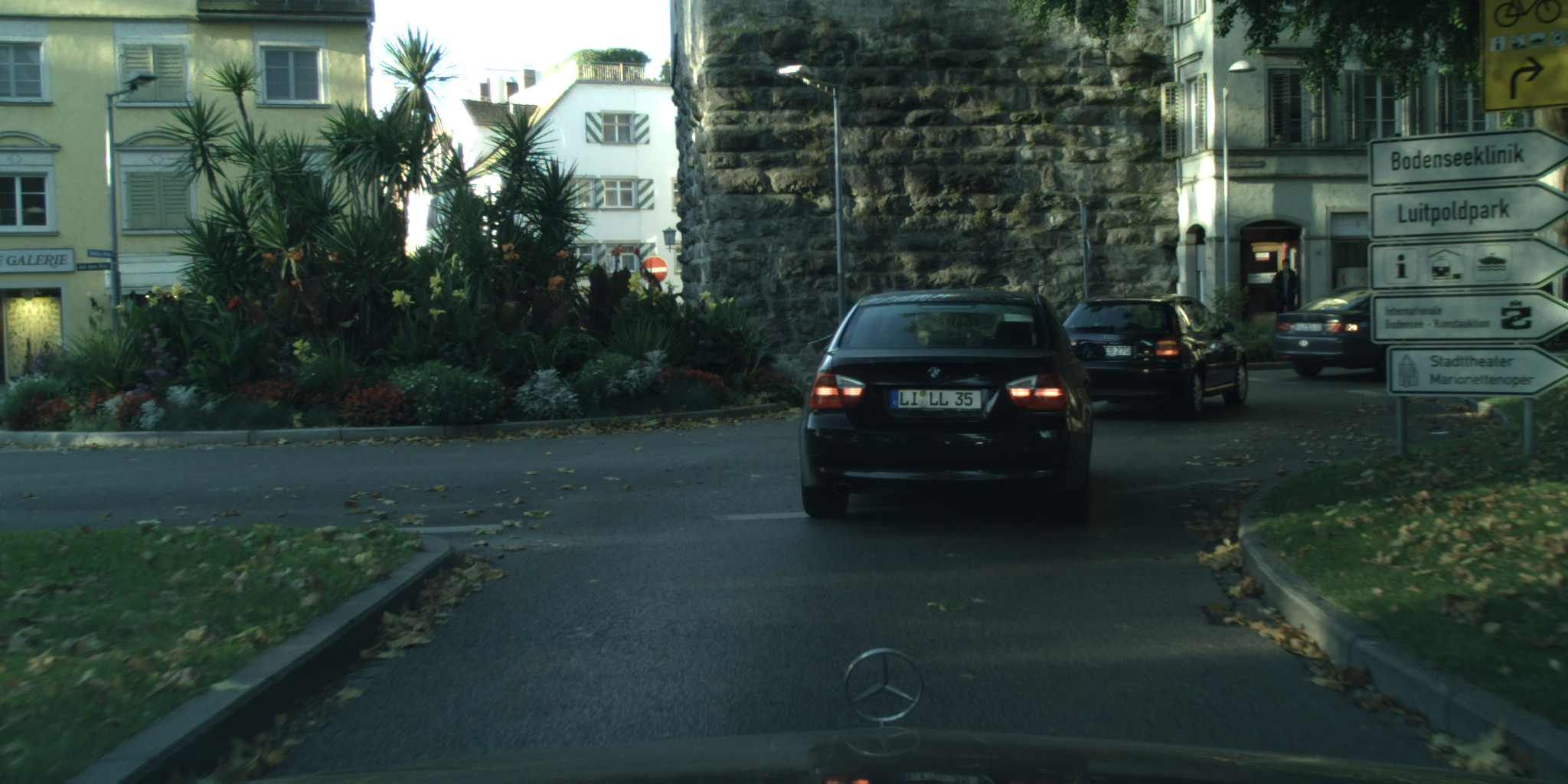} \\
				\includegraphics[width=\columnwidth]{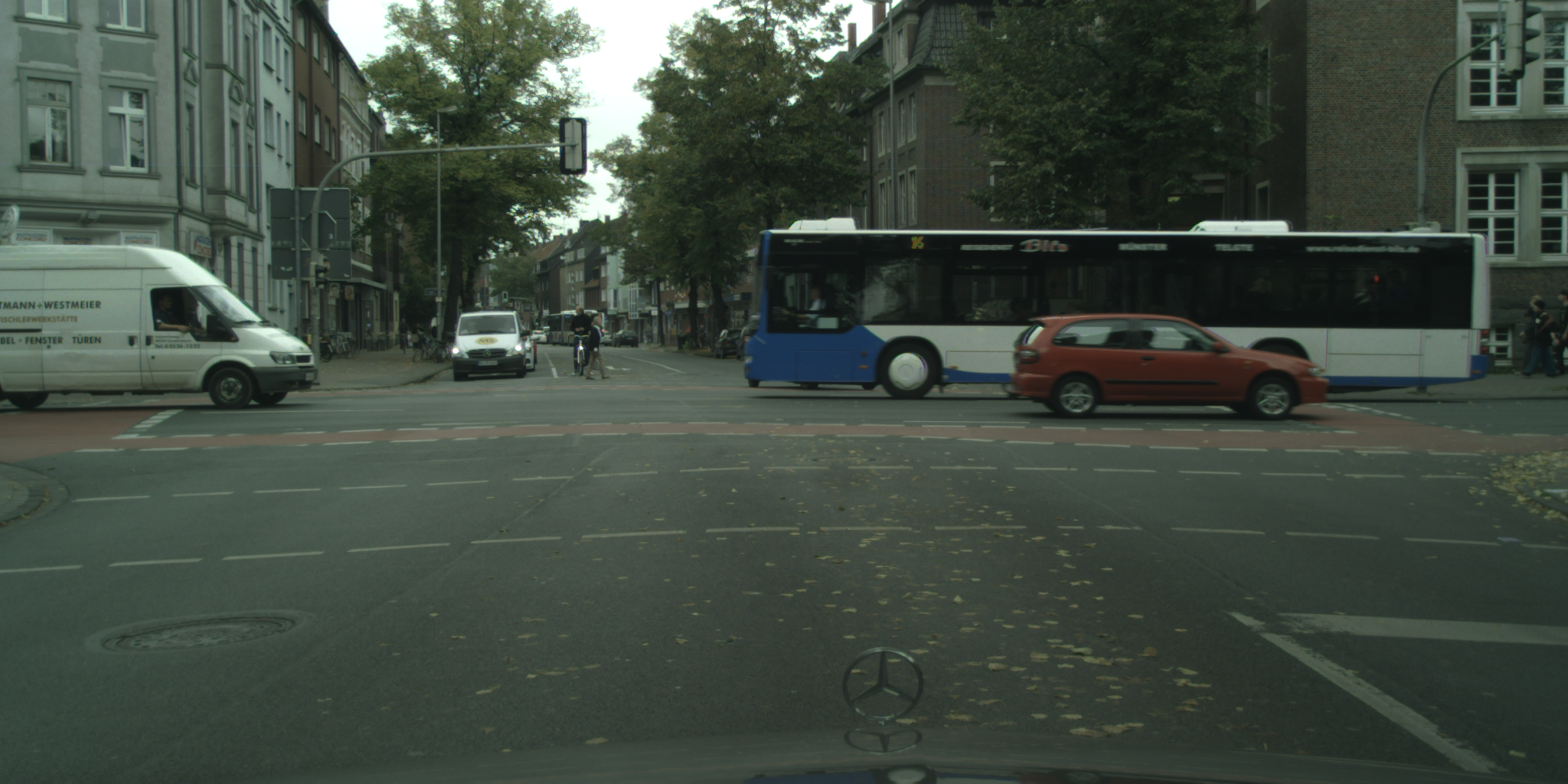}
			\end{minipage}
		}
		\subfloat[  GT]{
			\begin{minipage}[b]{0.30\columnwidth}
				\centering
				\includegraphics[width=\columnwidth]{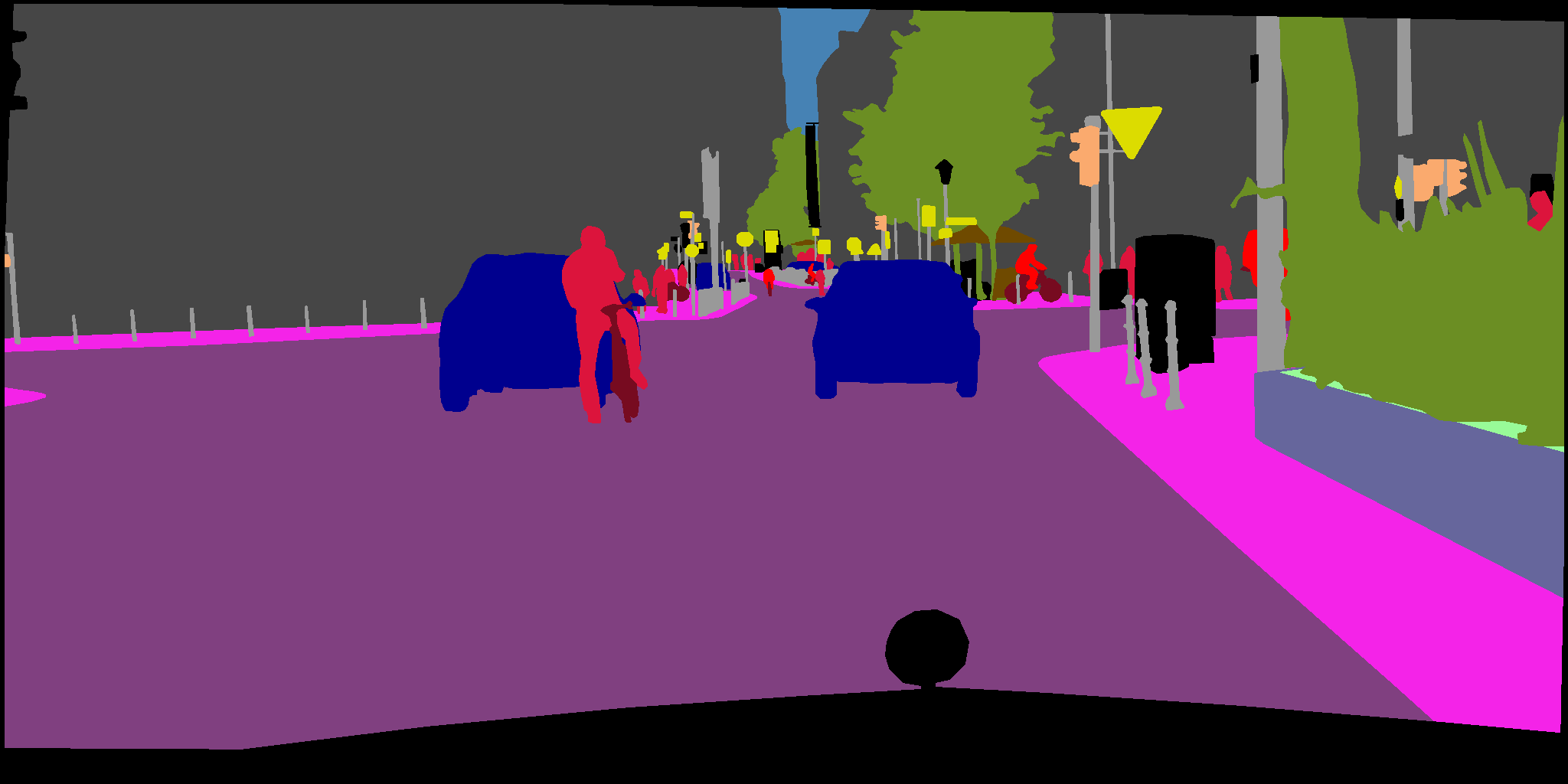} \\
				\includegraphics[width=\columnwidth]{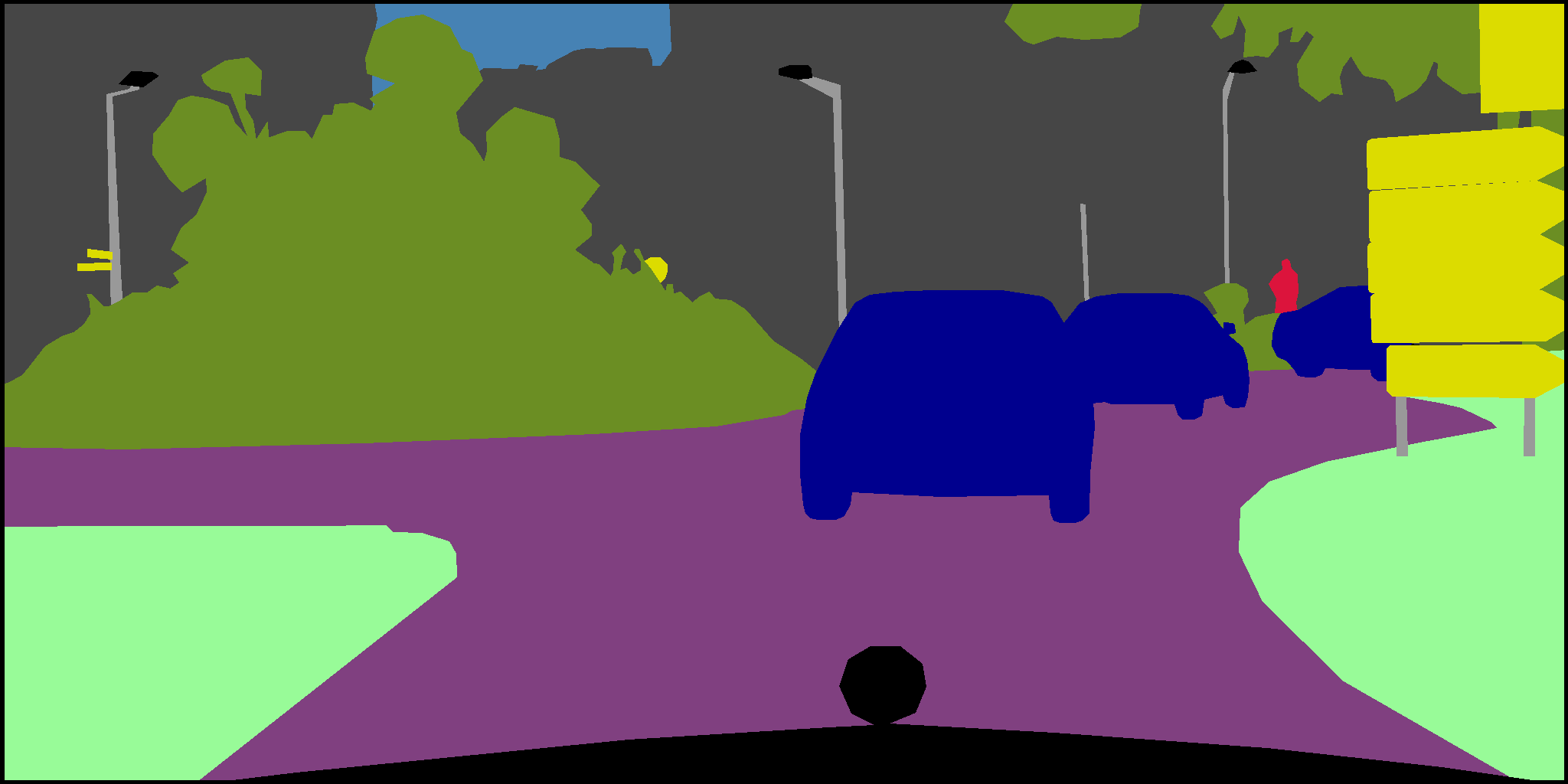} \\
				\includegraphics[width=\columnwidth]{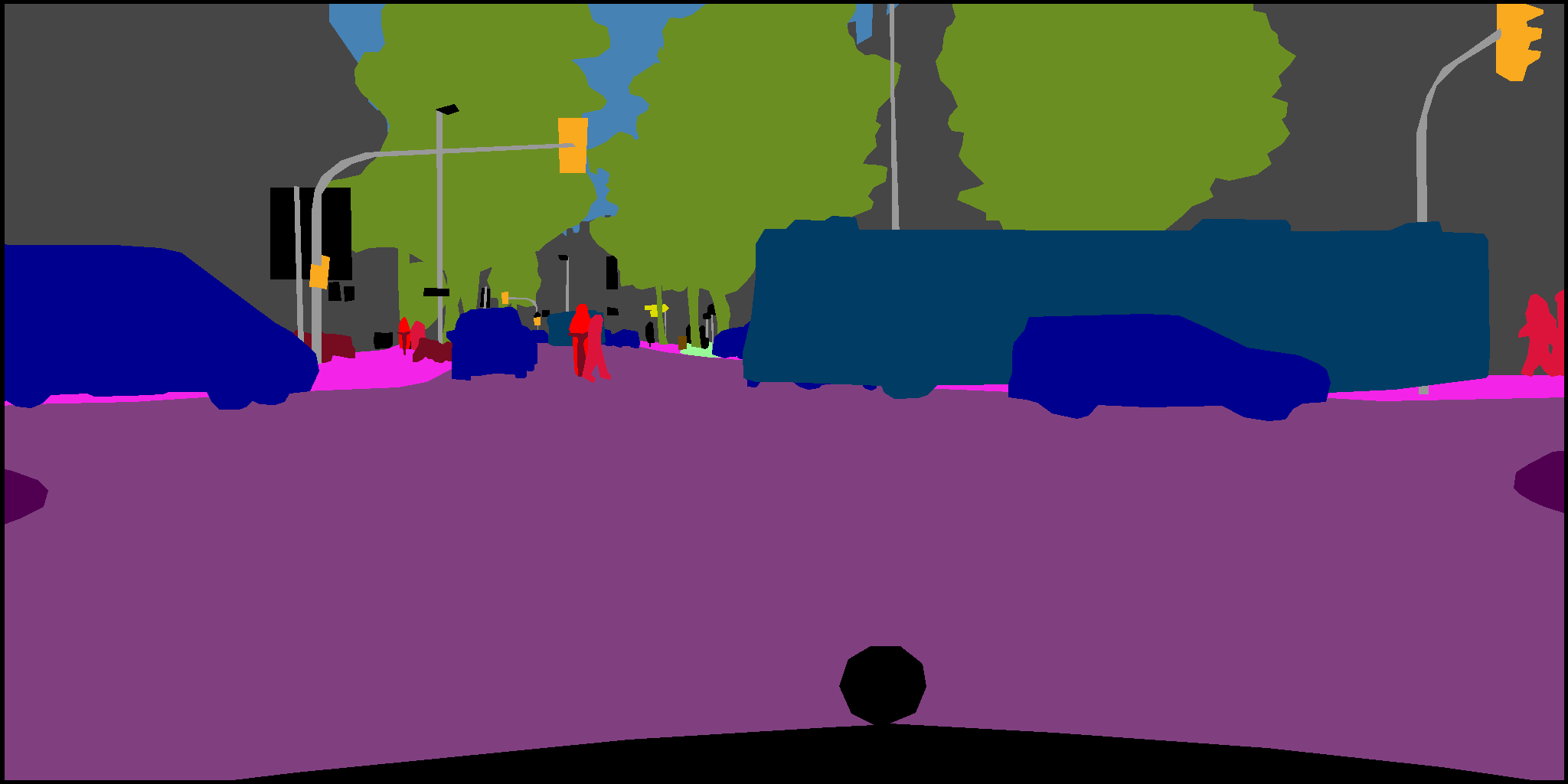}
			\end{minipage}
		}
		\subfloat[ Ours]{
			\begin{minipage}[b]{0.30\columnwidth}
				\centering
				\includegraphics[width=\columnwidth]{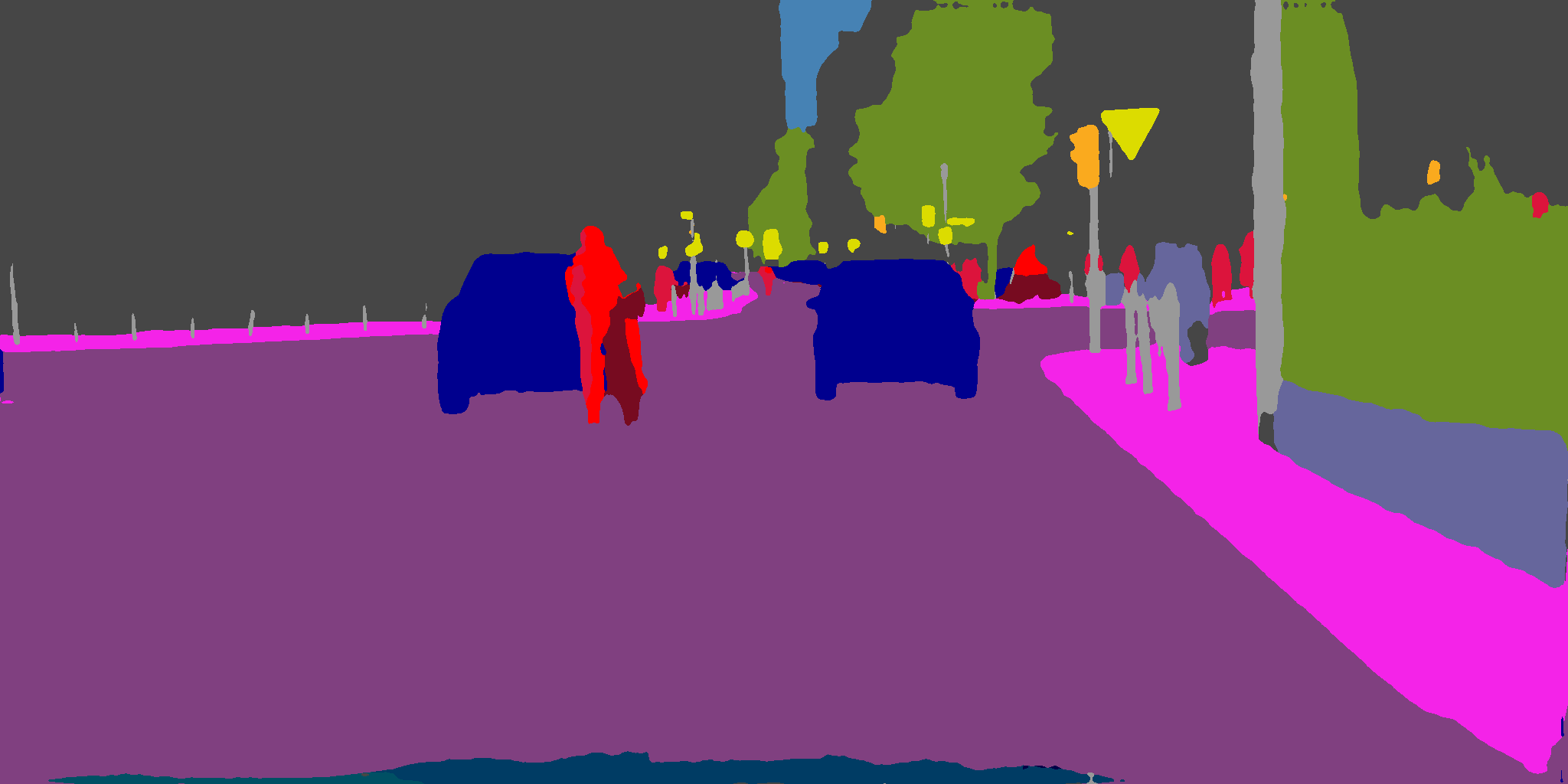} \\
				\includegraphics[width=\columnwidth]{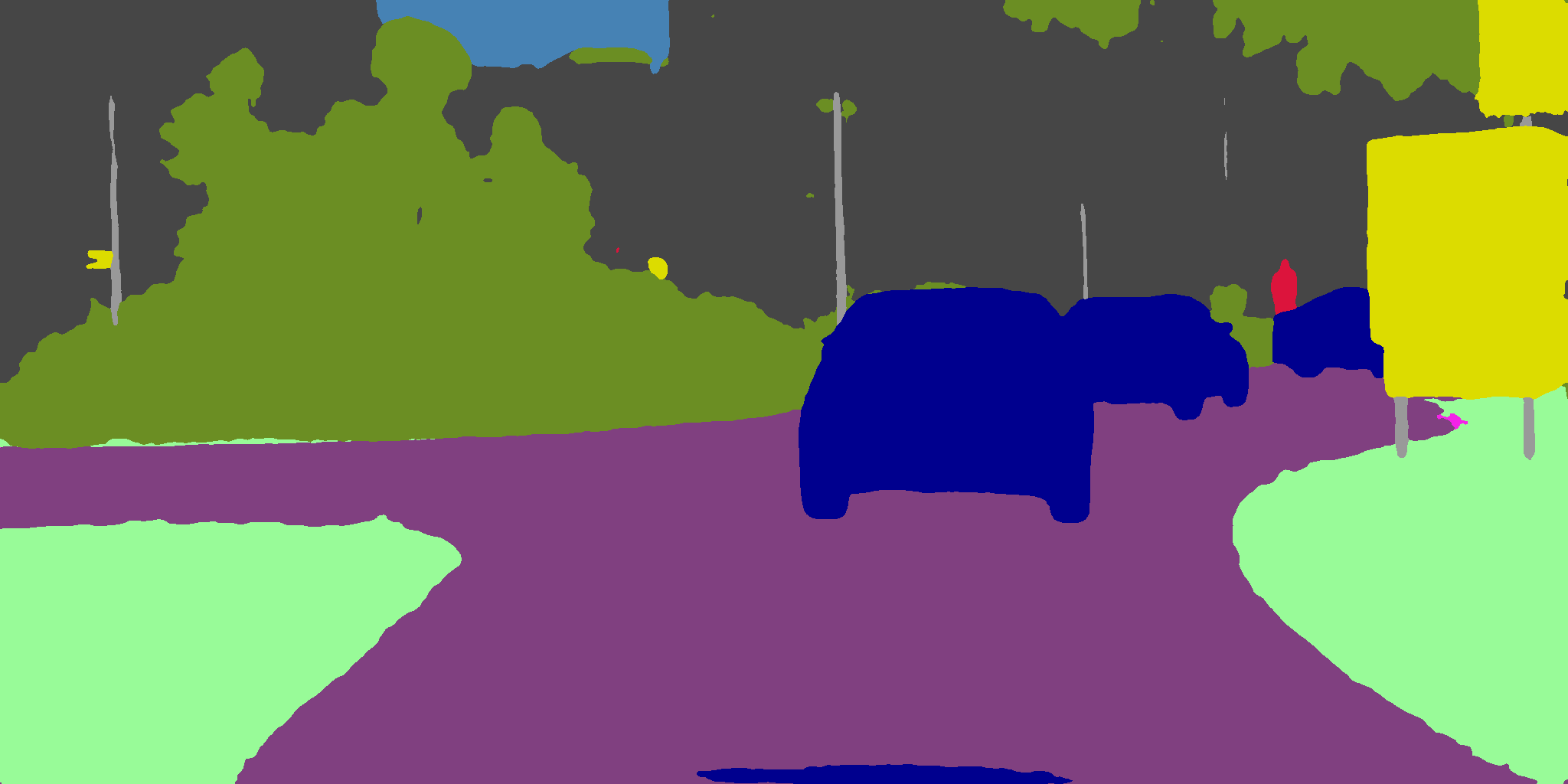} \\
				\includegraphics[width=\columnwidth]{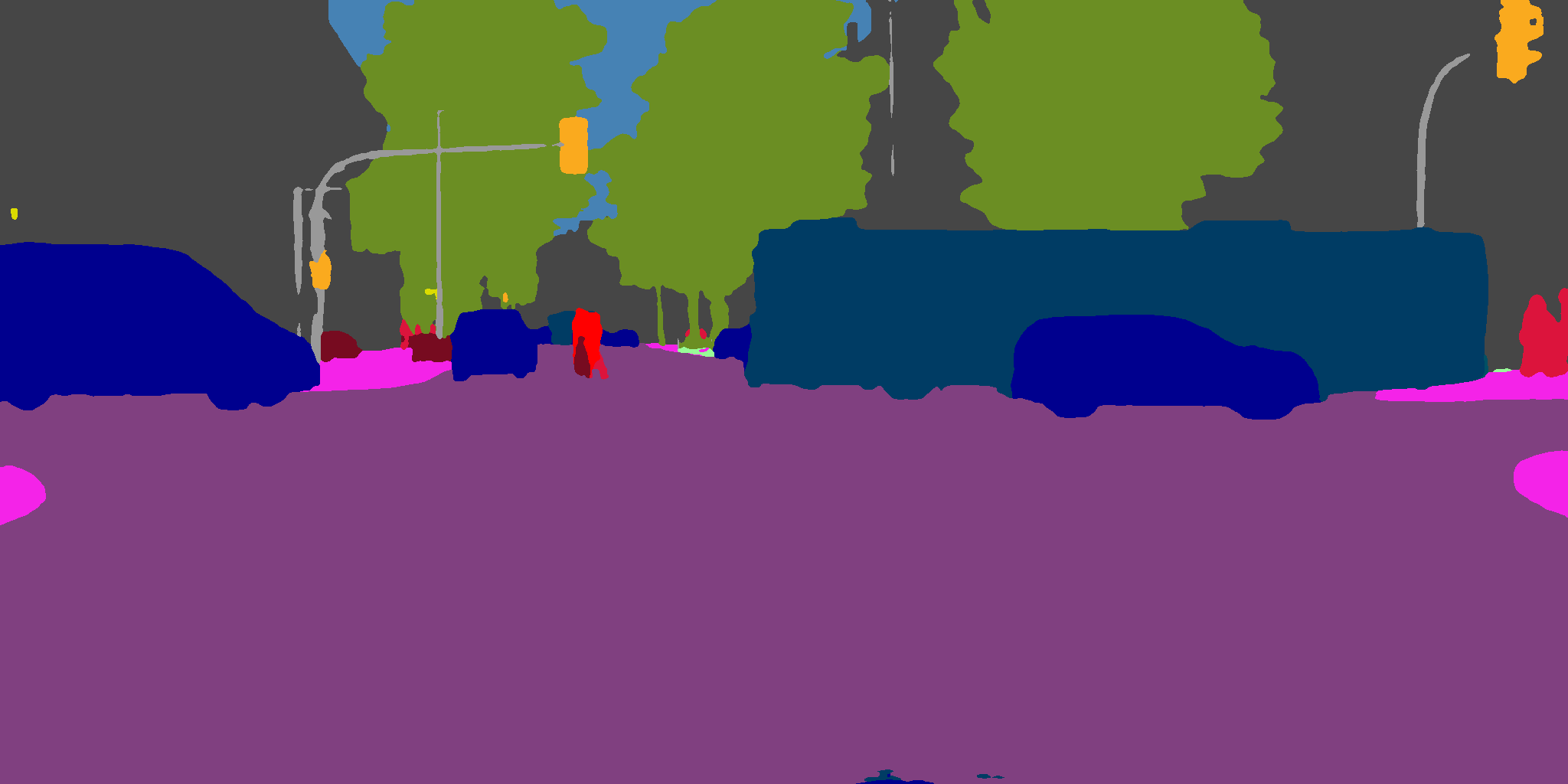}
			\end{minipage}
		}
	\end{center}
\vspace*{-2mm}
	\caption{Visualization results of our method on Cityscapes validation set.}
	\label{fig:city} 
\end{figure}

\subsubsection{More dataset}\label{S5.2.3}

We then evaluate our GDPP module on Cityscapes datasets. We use our GDC to replace the dilated convolution in the ASPP module of the original Deeplabv3. Here, the backbone network is Resnet101 \cite{He:2016ib} pre-trained on ImageNet dataset. The original DeeplabV3 achieves an mIoU of 80.1\% on the validation set. Our GDPP module can boost the mIoU to 80.6\%. The visualization results of our model can be found in Fig.~\ref{fig:city}. The comparison with more state-of-the-art models is shown in Table~\ref{tab:city}. It can be seen that our model achieves the highest mIoU. 

\begin{table}[h]
\vspace*{-1mm}
	\caption{Comparion of semantic segmentation performance on Cityscapes validation set.}
	\label{tab:city} 
\vspace*{-3mm}
	\begin{center}
		\begin{tabular}{p{0.3\columnwidth}|p{0.3\columnwidth}|p{0.2\columnwidth}{c}}
			\toprule
			\textbf{Methods} & \textbf{Backbone} & \textbf{mIoU (\%)} \\
			\midrule
			PSPNet \cite{2016arXiv161201105Z} & ResNet-101 & 77.8 \\
			PSANet \cite{zhao2018psanet} & ResNet-101 & 79.1 \\
			BiSeNet \cite{yu2018bisenet} & ResNet-101 & 80.3 \\
			DenseASPP \cite{yang2018denseaspp} & DenseNet-121 & 76.6 \\
			DeeplabV3 \cite{Chen:2018wj} & ResNet-101 & 80.1 \\
			\midrule
			Ours & ResNet-101 & \textbf{80.6} \\
			\bottomrule
		\end{tabular}
	\end{center}
\vspace*{-3mm}
\end{table}

\section{Discussion}\label{sec:dis}

As confirmed in Table~\ref{tab:single-performance}, our proposed GDC and the dilated convolution both outperform the normal convolution. Obviously, larger receptive field is beneficial for aggregating global contextual information. We now further discuss why the GDC outperforms the dilated convolution. Generally, the GDC can efficiently expand the receptive field without the need to stack many convolutional layers. The standard deviation $\Sigma$ of the Gaussian distribution controls the overall scale of the receptive field extending. As an illustration, we initialize a GDC kernel and sample 100 times. In Fig.~\ref{fig:rf}, we visualize the sample positions of the GDC kernel with two $\Sigma$ values at the central position $<x, y>$. Clearly, a larger $\Sigma$leads to a larger receptive field.
\begin{figure}[h]\setcounter{figure}{9}
	\vspace*{-6mm}
	\begin{center}
		\captionsetup[subfigure]{}
		\subfloat[]{
			\begin{minipage}[t]{0.47\columnwidth}
				\centering
				\includegraphics[width=1.5in,height=1.5in]{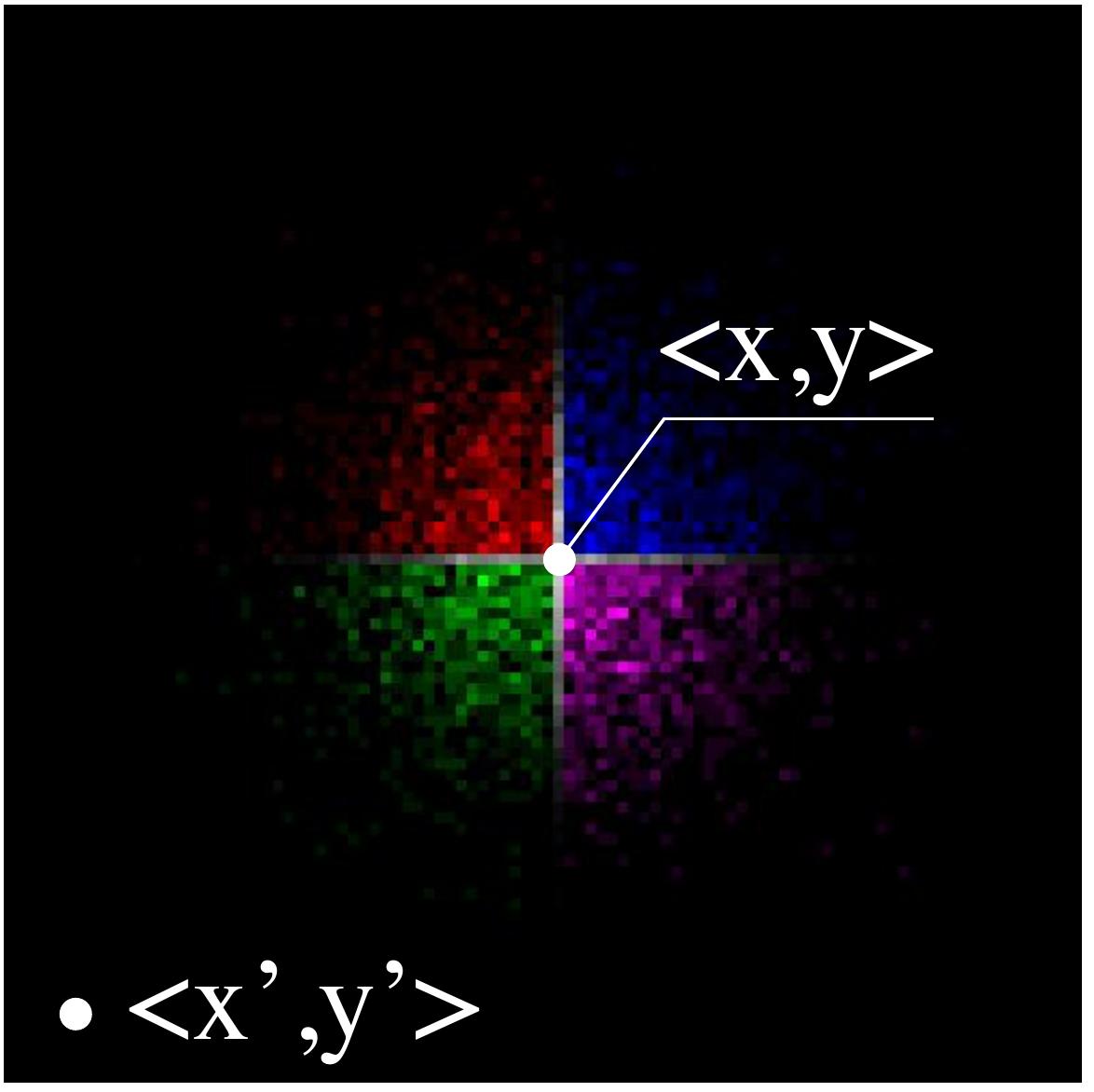}
			\end{minipage}
			\label{fig:rf1}
		}
		\subfloat[]{
			\begin{minipage}[t]{0.47\columnwidth}
				\centering
				\includegraphics[width=1.5in,height=1.5in]{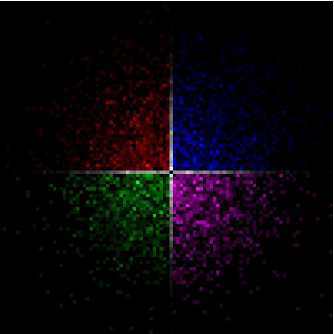}
			\end{minipage}
			\label{fig:rf2}
		}
	\end{center}
	\vspace*{-3mm}
	\caption{The sample points of the GDC kernel at position $<x, y>$ with (a) $\Sigma=0.1$, and (b) $\Sigma=0.15$.}
	\label{fig:rf} 
\end{figure}

\begin{figure*}[t]\setcounter{figure}{10}
	\begin{center}
		\begin{minipage}[t]{\textwidth}
			\centering
			\includegraphics[width=\textwidth]{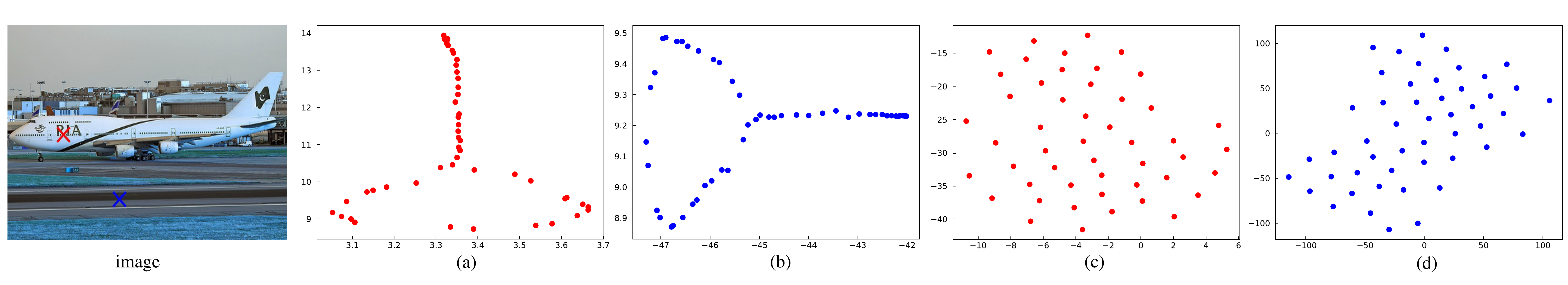}
		\end{minipage}
	\end{center}
\vspace*{-5mm}
	\caption{We use the tSNE to embed the feature vectors at the red and blue crosses of the image. Red and blue points are the embedded feature vectors of the red and blue cross positions, respectively. (a) and (b) are produced by the dilated convolution, while (c) and (d) by our GDC.}
	\label{disexampleimg} 
\vspace*{-2mm}
\end{figure*}

The core idea of the GDC however is to simulate the dynamic receptive field mechanism in the human being's visual system. Although the conventional dilated convolution can extend the receptive field, it can only use one or a group of fixed scales. This is totally different from the human visual system. By contrast, our GDC randomly selects different sampling positions to simulate this dynamic human receptive field. As illustrated in Table~\ref{tab:voc-result}, such a dynamic convolution can achieve better performance than the conventional dilated convolution. The reason is that our dynamic convolution introduces more randomness into the training process. Different sample positions can fuse richer features in various scales. The tail segmentation network trained with these richer features can be optimized to a more robust status. 

Moreover, vivid feature maps help to alleviate the problem of overfitting. The single image segmentation experiments of  Subsection~\ref{S5.1} offer the best evidence. Since there is only one image for optimizing the network, we need to prevent our lightweight segmentation network from overfitting into the scribble ground truth. Our dynamic convolution stochastically forms the convolution kernel of various spatial scales. In a way, it is equivalent to do some data augmentation in the feature space. For the sake of visual interpretation, we collect the feature vectors of the two positions, the red and blue crosses in the image of Fig.~\ref{disexampleimg} in 50 optimizing steps, and use tSNE \cite{maaten2008visualizing} to embed these feature vectors into two dimensions for visualizing. Visualizations of the embedded vectors are shown in the four scatter graphs in Fig.~\ref{disexampleimg}. Specifically, Figs.~\ref{disexampleimg}\,a and \ref{disexampleimg}\,b depict the embedded feature vectors of the dilated convolution, while Figs.~\ref{disexampleimg}\,c and \ref{disexampleimg}\,d show the feature vectors generated by our GDC. Obviously, the feature vectors produced by our GDC are more diverse than the dilated convolution. Thus, overfitting can be effectively alleviated.

\begin{table}[h]
  \vspace*{-4mm}
	\caption{An experiment on Pascal-VOC 2012 validation set.}
  \vspace*{-2mm}
	\label{tab:randomvsgaussian} 
	\begin{center}
		\begin{tabular}{p{0.22\columnwidth}p{0.32\columnwidth}|p{0.16\columnwidth}}
			\toprule
			\textbf{Template} & \textbf{Setting} & \textbf{mIoU (\%)}\\
			\midrule
			Baseline & normal conv $3\times3$ & 68.94\\
			GDC & $\Sigma=0.2$ & 74.06 \\
			Random & - &  68.10\\
			\bottomrule
		\end{tabular}
	\end{center}
\end{table}

The final discussion is why we sample the offset values from the Gaussian distribution. We believe that Gaussian distribution can simulate the correlation between feature vectors. For example, in Fig.~\ref{fig:rf1}, the feature vector located at $<x',y'>$ has few relation with the feature vector located at $<x, y>$. If we aggregate the information at $<x',y'>$ into $<x, y>$, the feature map will become turbid. We design an simple experiment to illustrate this point. We replace the GDC in the lightweight segmentation network with a random dynamic convolution. The offset values of this random dynamic convolution are sampled from a uniform distribution. The result is shown in Table~\ref{tab:randomvsgaussian}. Observe that the random dynamic convolution achieves an mIoU of 68.10\% which is even lower than the baseline.

\section{Conclusions}\label{S7}

In this paper, we have propose a novel Gaussian Dynamic Convolutions for the fast and effective single-image segmentation task. The proposed GDC can dynamically change its receptive field by sampling different offset values from a Gaussian distribution. This GDC can not only be employed for the single-image segmentation with scribbles but also be implemented for the common semantic segmentation network with a Gaussian Dynamic Pyramid Pooling. Our experiments have demonstrated that the Gaussian dynamic convolution achieves better performance on the image segmentation than other existing forms of convolution, including dilated convolution and deformable convolution. In addition, we believe that the novel GDC can help other computer vision tasks, such as classification and detection, which will be our future work.


\bibliography{citebib}
\bibliographystyle{IEEEtran}

\end{document}